\documentclass[journal]{IEEEtran} 
\usepackage{amsmath,amsfonts}
\usepackage{algorithmic}
\usepackage{algorithm}
\usepackage{array}
\usepackage[caption=false,font=normalsize,labelfont=sf,textfont=sf]{subfig}
\usepackage{textcomp}
\usepackage{stfloats}
\usepackage{url}
\usepackage{verbatim}
\usepackage{graphicx}
\usepackage{cite}
\usepackage{xcolor}
\usepackage{glossaries}
\usepackage{siunitx}
\usepackage{longtable}
\usepackage{booktabs}
\usepackage{array}
\usepackage{pifont}

\newacronym{mt}{MT}{mechanical thrombectomy}
\newacronym{threed}{3D}{three-dimensional}
\newacronym{ai}{AI}{artificial intelligence}

\newacronym{Hz}{Hz}{hertz}
\newacronym{sla}{SLA}{stereolithography}
\newacronym{cad}{CAD}{computer-aided design}
\newacronym{cv}{CV}{coefficient of variation}
\newacronym{stl}{STL}{stereolithography}
\newacronym{kch}{KCH}{King’s College Hospital}
\newacronym{sie}{SIE}{Surgical \& Interventional Engineering}
\newacronym{ica}{ICA}{internal carotid artery}
\newacronym{bca}{BCA}{brachiocephalic artery}
\newacronym{cca}{CCA}{common carotid artery}
\newacronym{rica}{RICA}{right internal carotid artery}
\newacronym{lcca}{LCCA}{left common carotid artery}
\newacronym{reca}{RECA}{right external carotid artery}
\newacronym{rcca}{RCCA}{right common carotid artery}
\newacronym{rsa}{RSA}{right subclavian artery}

\newacronym{lvo}{LVO}{large vessel occlusion}
\newacronym{ais}{AIS}{acute ischemic stroke}
\newacronym{mca}{MCA}{middle cerebral artery}

\newacronym{rc}{RC}{robotic controller}
\newacronym{iqr}{IQR}{interquartile range}

\newacronym{sac}{SAC}{soft actor-critic}
\newacronym{rl}{RL}{reinforcement learning}
\newacronym{mlp}{MLP}{multi-layer perceptron}
\newacronym{lstm}{LSTM}{long short-term memory}
\newacronym{t2t}{T2T}{time-to-target}
\newacronym{std}{STD}{standard deviation}

\newacronym[
    plural=Copilot Assistance Modes,
    longplural={Copilot Assistance Modes}
]{cam}{CAM}{Copilot Assistance Mode}
\newacronym{pla}{PLA}{Polylactic Acid}

\newacronym{or}{OR}{operating room}
\newacronym{cta}{CTA}{computed tomography angiography}
\newacronym{tlx}{NASA-TLX}{NASA Task Load Index}
\newacronym{sus}{SUS}{System Usability Scale}
\newacronym{ucid}{UCID}{user control input device}
\newacronym{ui}{UI}{user interface}

\newcommand{\agent}{AI agent}
\newcommand{\phantommodel}{phantom model}

\newcommand{\copilotPushButton}{Copilot push-button}
\newcommand{\AIdrivenmodes}{AI-driven modes}
\newcommand{\AIDrivenModes}{AI-Driven Modes}

\newcommand{\CamsWrittenOut}{Copilot Assistance Modes (CAMs)}
\newcommand{\camnull}{Manual (0)}
\newcommand{\camone}{Direct Robot Control (1)}
\newcommand{\camtwo}{AI Advice (2)}
\newcommand{\camthree}{AI Assistance (3)}
    \newcommand{\camthreea}{AI Translation (3a)}
    \newcommand{\camthreeb}{AI Rotation (3b)}
\newcommand{\camfour}{Full Copilot Assistance (4)}
    \newcommand{\camfoura}{Full Copilot--Predefined Sections (4a)}
    \newcommand{\camfourb}{Full Copilot--Operator-Defined Sections (4b)}
\newcommand{\camfive}{Fully Autonomous Navigation (5)}

\newcommand{\invitrodash}{\textit{in-vitro}}
\newcommand{\invivodash}{\textit{in-vivo}}
\newcommand{\insilico}{\textit{in silico}}
\newcommand{\invitro}{\textit{in vitro}}

\newcommand{\tableref}{TABLE}
\newcommand{\supmaterial}{Supplementary Material}
\newcommand{\figref}{Figure}
\newcommand{\secref}{Section}

\newcommand{\colorgradientWhiteGreyBlue}{The blue color gradient indicates increasing levels of autonomy, ranging from white (manual) over grey (teleoperation) to dark blue (highest autonomy).}
\newcommand{\colorgradientGreyBlue}{The blue color gradient indicates increasing levels of autonomy, ranging from grey (teleoperation) to dark blue (highest autonomy).}

\begin{document}
\title{Clinician-AI Collaboration in Endovascular Neurointerventions: \textit{In-Vitro} User Study of Copilot Assistance Modes for Robotic Navigation Across the Autonomy Spectrum}

\author{
Nikola~Fischer\textsuperscript{1},
Harry~Robertshaw\textsuperscript{1},
S.M.Hadi~Sadati\textsuperscript{1,2},
Xingyu~Chen\textsuperscript{1},
Weiyuan~Deng\textsuperscript{1},
Benjamin~Jackson\textsuperscript{1},
Michael~Adofo~Kwakye\textsuperscript{1},
Sara~Sciacca\textsuperscript{3},
Naga~Kandasamy\textsuperscript{3},
Parthiban~Balasundaram\textsuperscript{3},
Sahan~Guruge\textsuperscript{3},
Charles~Cummins\textsuperscript{3},
Joe~Leyon\textsuperscript{3},
Nasr~Abdelsalam\textsuperscript{4},
Matthew~Benger\textsuperscript{5},
Andrew~Falzon\textsuperscript{5},
Alejandro~Granados\textsuperscript{1},
Christos~Bergeles\textsuperscript{1\textdagger},
and Thomas~Booth\textsuperscript{1,3*\textdagger}

\thanks{
\hspace*{-\footnotesep}
\parbox{\dimexpr\columnwidth-0.5em\relax}{%
\footnotesize
\textsuperscript{1}School of Biomedical Engineering \& Imaging Sciences, King's College London, UK; 
\textsuperscript{2}School of Engineering and Materials Science, Queen Mary University of London, UK;  
\textsuperscript{3}Department of Neuroradiology, King’s College Hospital NHS Foundation Trust, London, UK; 
\textsuperscript{4}The Walton Centre NHS Foundation Trust, Liverpool, UK; 
\textsuperscript{5}St George's University Hospitals NHS Foundation Trust, London, UK\\
*Corresponding author: thomas.booth@kcl.ac.uk\\
\textsuperscript{\textdagger}These authors contributed equally as senior authors.\\ 
This work was supported by the Medical Research Council (MRC) Impact Acceleration Account (IAA) 2021 at King's College (Grant No. MR/X502923/1), by EPSRC [VITRRO / EP/Y024281/1 - ERC Proof of Concept] under the Horizon Europe Guarantee Extension, the King's-funded Centre for Doctoral Training on Surgical \& Interventional Engineering, and the King's-China Scholarship Council [CSC No.: 202408330072]. Work in the Surgical \& Interventional Engineering Validation Suite (mock OR) was supported by core funding from the Wellcome/EPSRC Centre for Medical Engineering [WT203148/Z/16/Z], a multi-user equipment grant for post-mortem evaluation of medical devices from Wellcome [218286/Z/19/Z], and the Wolfson Foundation [PR/ylr/md/21896].
}
}
}

\maketitle

\begin{abstract}
Mechanical thrombectomy (MT) is an effective treatment for acute ischemic stroke, yet access remains limited by the availability of specialist neurointerventional expertise. Telerobotic and AI-assisted systems may help address this challenge, but it remains unclear how clinicians interact with different levels of autonomy and how autonomous navigation systems perform in clinically realistic environments.
This work introduces \CamsWrittenOut{}, a framework for clinician-AI collaboration in robotic endovascular navigation spanning the autonomy spectrum from manual operation to full autonomy. In the first \invitrodash{} user study systematically comparing clinician-AI collaboration strategies, 11 expert and two student participants evaluated six navigation modes in a patient-specific vascular phantom under fluoroscopic guidance. Objective performance metrics were combined with subjective assessments, including workload, usability, and operator preference.
Autonomous performance was additionally evaluated both \insilico{} and \invitro{} to investigate long-horizon sim-to-real transfer.
Hands-on manual navigation achieved the strongest overall performance (100\% success).
While fully autonomous navigation achieved 81\% success in simulation, performance dropped to 23\% \invitro{}. In contrast, several clinician-AI shared-autonomy modes achieved success rates of 92–97\%, approximately four times higher than full autonomy and comparable to direct robotic control (95\%), potentially demonstrating mitigation to sim-to-real domain shift when using imperfect AI through clinician-AI collaboration.
Among the evaluated AI-driven approaches, user-directed task delegation and rotational shared control demonstrated the most favorable balance between performance and clinician acceptance. 
These findings plausibly suggest that, for robotic neurointerventional navigation, effective clinician-AI collaboration and allocation of control authority may currently be more important for reliable performance than increasing autonomy alone.
\end{abstract}

\begin{IEEEkeywords}
human-AI, human--machine, shared autonomy, autonomous navigation, zero-shot sim-to-real transfer
\end{IEEEkeywords}

\section{Introduction}\label{sec_introduction}
\IEEEPARstart{S}{troke} is one of the leading causes of death and long-term disability worldwide, with a substantial socioeconomic burden on healthcare systems \cite{feigin2021global}. \Gls{ais} remains a major global health burden, with approximately 12 million new cases and 7 million deaths annually~\cite{Feigin2025}. \Gls{lvo} strokes are particularly time critical, as reduced cerebral perfusion in a large territory accelerates considerable neuronal loss, making rapid recanalization essential for preserving neurological function and improving clinical outcomes. \Gls{mt} has emerged as the reference standard treatment for eligible patients, significantly improving functional outcomes and reducing mortality compared to medical therapy alone \cite{bendszus2023endovascular, goyal2016, nogueira2018}.

\begin{figure}[t]
    \centering
    \includegraphics[width=.9\linewidth]{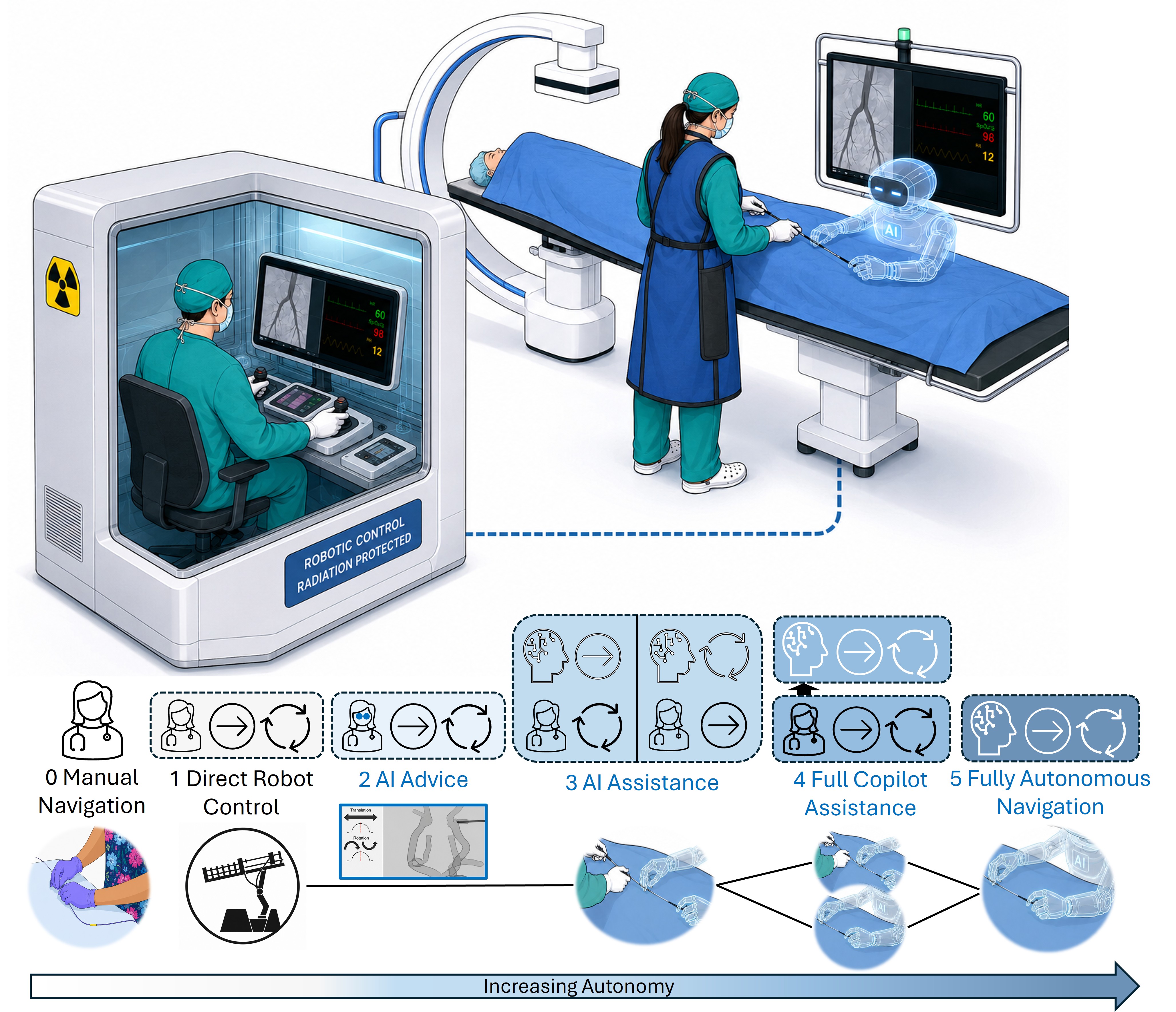}
    \caption{\textbf{Clinician-AI Collaboration for Neurointerventions:} Introducing the \CamsWrittenOut{} including the conventional \camnull{} and \camone{} as well as the \AIdrivenmodes{} (blue) \camtwo{}, \camthree{}, \camfour{}, and \camfive{}. \colorgradientWhiteGreyBlue{}
    }
    \label{fig:introduction_visual_abstract_cam_modes}
\end{figure}
Despite its clinical effectiveness, access to \gls{mt} remains highly unequal. Because treatment success is critically dependent on minimizing the time from symptom onset to reperfusion, the limited availability of specialized neurointerventional centers and trained operators prevents many eligible patients from receiving timely treatment~\cite{Kaiz_2023_access_MT}, thereby reducing the likelihood of favorable clinical outcomes~\cite{time_is_brain}.
The workforce shortage~\cite{RCR2025WorkforceCensus} is further exacerbated by the technical complexity of endovascular interventions, in which operators must navigate instruments (i.e., guidewires and catheters) through highly tortuous vasculature while achieving predictable control of the distal instrument tip through indirect manipulation of the two degrees of freedom (DoF) rotation $\varphi$ and translation $d$ at the proximal end. As a result, accurate navigation places considerable demands on clinician dexterity, hand-eye coordination, and procedural experience.
To facilitate instrument handling, robotic (neuro)endovascular systems are emerging~\cite{crinnion2022robotics}. These systems enable precise teleoperated manipulation of endovascular instruments, potentially also allowing expert clinicians to perform procedures remotely from centralized stroke centers~\cite{Shen2024_treatment_stroke_centers}. This paradigm may reduce geographic disparities by concentrating expertise while extending procedural reach to peripheral medical centers. In addition, robotic actuation may reduce radiation exposure and improve operator ergonomics~\cite{klein2009, madder2017}, even for clinicians performing procedures locally. 

Beyond pure teleoperation, robotic systems also create opportunities for enhancement with \gls{ai} of the challenging navigation from the vascular access site to the intracranial occlusion. As this navigation requires operators to achieve predictable instrument tip behavior through indirect control and continuous adaptation, it remains a major procedural bottleneck and a potential source of serious complications (risks), including procedure-related vessel perforation (1\%), dissection (2\%), and distal embolization of the thrombus (9\%)~\cite{berkhemer20215}.

Outside of neurointervention, early efforts focused on automating constrained procedural subtasks~\cite{robertshaw_2023_AI_navigation_review}. Chen et al. presented an autonomous vascular access system using deep-learning-based robotic guidance, demonstrating the feasibility of \gls{ai}-assisted vascular procedures~\cite{Chen2020}. Building on this for neurointervention, Abah et al. proposed a semi-autonomous catheter steering framework based on path-planar algorithms and preplanned optimal catheter configurations~\cite{Abah2024}.
Yet, maintaining robust performance over long-horizon navigation tasks remains a key challenge in autonomous endovascular navigation, with recent work highlighting limitations in the generalization of \gls{rl} policies as task complexity and navigation duration increase \cite{Robertshaw2026TowardHM_MARL}.

Recent developments in \gls{ai}-driven robotic control systems, also described as agentic control systems or \gls{ai} copilots~\cite{Wang2026_copilot}, have increasingly shifted attention toward collaborative paradigms that combine \gls{ai} capabilities with clinician expertise, reflecting the emerging vision of surgical teams in which clinicians, \gls{ai} systems, and robotic platforms operate as collaborative partners~\cite{granados2026}. However, effective integration of human-\gls{ai} collaboration remains an open challenge. 
In the endovascular domain, Yan et al. introduced a machine-learning-based human–machine collaborative control strategy for guidewire insertion that dynamically adapts to varying surgical conditions and incorporates a trust evaluation model to improve procedural safety and efficiency~\cite{Yan2022}. 
Such systems are increasingly viewed as shared autonomy agents designed to enhance procedural precision, reduce cognitive workload, and improve navigation efficiency.
This way, \gls{ai}-assisted robotic systems can help broaden access to care by supporting less experienced operators (i.e. novices) or even clinicians without specialized training during demanding procedural steps. 

Despite these advances shown in \cite{Chen2020, Abah2024, Robertshaw2026TowardHM_MARL, granados2026, Yan2022, Xian2023, Zhang2024, robertshaw_2023_AI_navigation_review}, existing \gls{ai} research for robotic endovascular use cases such as \gls{mt}, remains primarily focused on the development of increasingly autonomous navigation systems and improvements in navigation performance. In contrast, considerably less attention has been paid to understanding how clinicians and AI systems should collaborate during robotic endovascular procedures. As a result, it remains largely unclear how clinician-\gls{ai} teams perform in robotic endovascular navigation settings, how operators prefer to interact with \gls{ai} copilots in shared-control scenarios, and how different distributions of control authority influence navigation performance, workload, usability, and clinician perception. 

Addressing these questions is critical for the clinical translation of AI-assisted robotic intervention, where successful adoption will likely depend not only on autonomous capabilities, but also on effective mechanisms for clinician-\gls{ai} collaboration. Accordingly, this study investigates how different levels of AI-driven autonomy~\cite{yang2017_autonomy} influence objective navigation performance, operator workload, usability, and clinician perception during robotic endovascular navigation for \gls{mt} (\figref{}~\ref{fig:introduction_visual_abstract_cam_modes}).

Specifically, this work makes the following contributions:
\begin{itemize}
    \item We introduce the \CamsWrittenOut{}, a framework for systematically investigating clinician-AI collaboration in robotic neurointerventional navigation. We instantiate this framework through multiple assistance paradigms representing guidance, shared control, delegated control, and fully autonomous navigation. 
    \item We developed a comprehensive \invitrodash{} evaluation framework that combines objective navigation outcomes with subjective assessments of workload, usability, and clinician perception for the evaluation of AI-assisted robotic interventions. 
    \item We present, to our knowledge, the first \invitrodash{} user study with clinical experts systematically comparing clinician-AI collaboration modes for robotic neurointerventional navigation under fluoroscopic guidance. 
   \item We provide one of the first empirical investigations into how different allocations of control authority between clinicians and AI systems influence subjective and objective metrics, while also highlighting challenges associated with long-horizon autonomous navigation and zero-shot sim-to-real transfer.
\end{itemize}

\section{Methods} \label{sec_Methods}
The methodology is structured to progressively build the experimental framework used to investigate clinician-AI collaboration in robotic neurointerventional navigation. 
First, we define the navigation task and describe the \invitrodash{} experimental platform, including the patient-specific vascular phantom, robotic manipulator, and user control interface. 
Next, we present the AI navigation agent that provides autonomous navigation capabilities. Building on these components, we introduce the \glspl{cam}, a framework for systematically studying different distributions of control authority between clinician and AI. 
Finally, we describe the user study design, outcome measures, and statistical analysis used to evaluate the proposed collaboration modes.

\subsection{Endovascular Navigation Task}\label{subsec_method_navigation_task}
Endovascular navigation for a \gls{mt} procedure involving a thrombus located in the distal cerebral vasculature typically begins at the femoral artery. From there, the device is advanced through the descending aorta to the aortic arch, followed by selective catheterization of the \gls{rcca} or the \gls{lcca} respectively, followed by subsequent progression into the internal carotid artery, and further advancement toward the intracranial target vessels -- commonly the M1 segment of the \gls{mca}.

\begin{figure}
    \centering
    \includegraphics[width=1\linewidth]{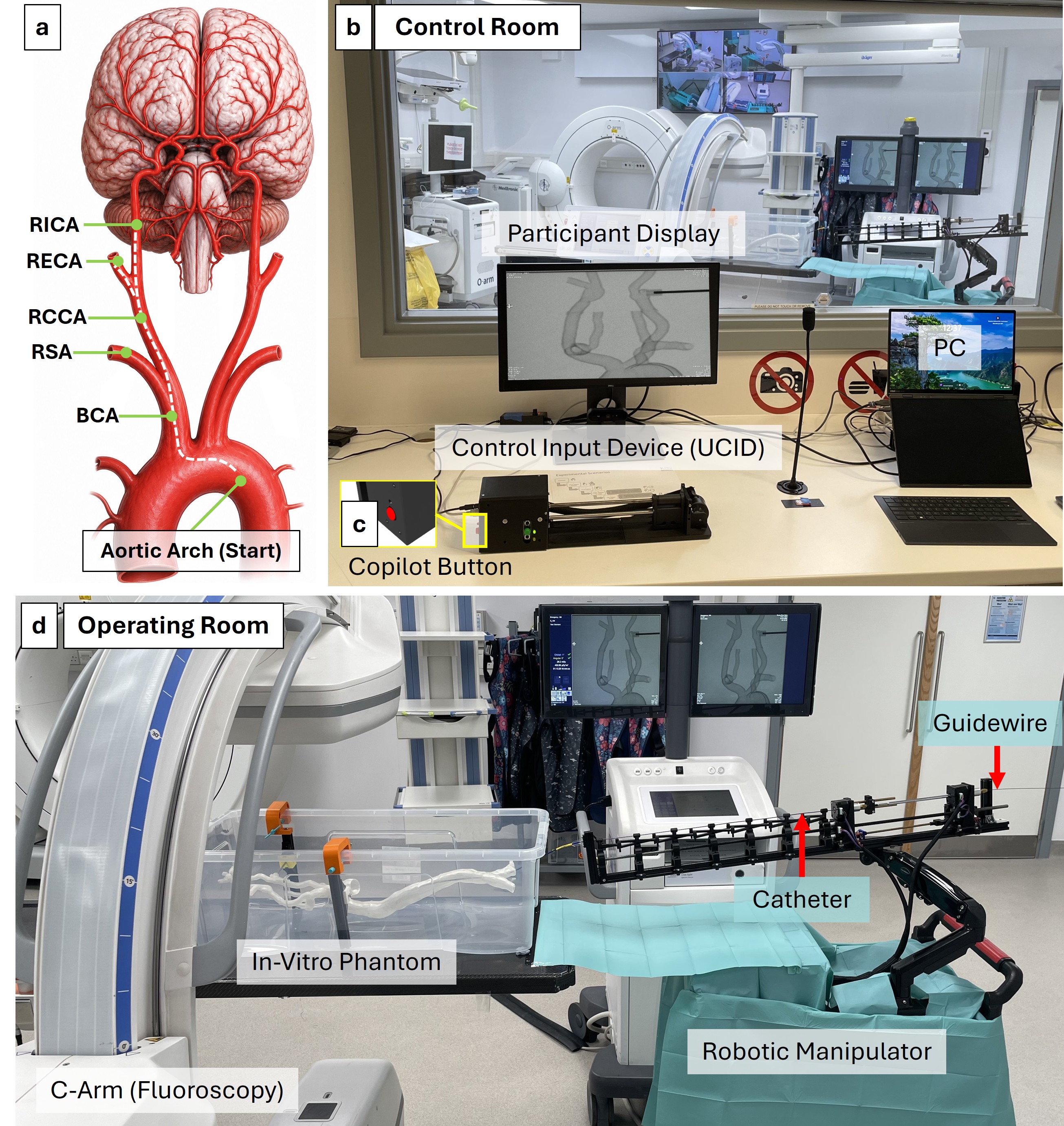}
    \caption{\textbf{Experimental Setup:} 
    (a) Navigation path (white dashed line) from the descending aorta through the aortic arch and brachiocephalic artery (BCA) into the right common carotid artery (RCCA), bypassing the right subclavian artery (RSA), and proceeding to either the right internal carotid artery (RICA) or the right external carotid artery (RECA).
    (b) Control room from which the participants are teleoperating the robotic manipulator utilizing the User Control Input Device (UCID)~\cite{jackson2026}. 
    (c) A red \copilotPushButton{} functions as a dead-man switch, activating autonomous navigation.
    (d) Mock operating room with the robotic manipulator for catheter and guidewire~\cite{sadati2025autonomous}, navigating within the \invitrodash{} phantom under fluoroscopy imaging.
    }
    \label{fig:method_experimental_setup}
\end{figure}

The navigation task in this work was defined as advancing from the aortic arch to the target vessels \gls{rica} and \gls{reca}, respectively (\figref{}~\ref{fig:method_experimental_setup}a). This segment is particularly relevant due to the anatomical complexity of the aortic arch branching with the need for precise instrument steering to ensure successful supra-aortic vessel selection.

Compared with the majority of autonomous endovascular navigation studies which are typically focused on tasks around the aortic arch~\cite{robertshaw_2023_AI_navigation_review}, the investigated task represents an extended navigation horizon in both traversed distance and procedure duration.
The estimated navigation distance for solely the aortic arch, from starting point to the \gls{bca} is $\approx$250\,mm, while our extended navigation toward \gls{rica} and \gls{reca} require a distance of $\approx$425\,mm.

We used a patient-specific anatomical model reconstructed from clinical imaging data, namely a neck \gls{cta} scan extending from the aortic arch to the cerebral arteries and a body \gls{cta} scan encompassing the thoracic and abdominal vasculature, including the femoral arteries, descending aorta, and aortic arch (U.K. Research Ethics 24/LO/0057).
The \invitrodash{} test anatomy comprised a type I aortic arch without a bovine arch configuration, a reverse-curved \gls{rica}, and a high-riding \gls{bca} with a take-off angle of \SI{141}{\degree}. Recent evidence suggests that these anatomical characteristics are associated with reduced navigation success rates, with the reverse \gls{rica} curvature further contributing to longer procedure times \cite{ruwu_vasc_geometry2026}.
Failure to negotiate these challenges may result in navigation errors and prolonged procedure times. Together, these factors make the navigation task sufficiently rigorous for evaluating clinician-AI collaboration.

\subsection{Experimental Setup for in-vitro Experiments}\label{subsec_method_experimental_setup}
For the \invitrodash{} testbed, we fabricated a 3D-printed vascular \phantommodel{} from translucent \gls{pla} based on the aforementioned patient anatomy (\secref{}~\ref{subsec_method_navigation_task}).
For endovascular instrumentation, a guidewire (Glidewire, 180 cm, 0.035”, Terumo, Tokyo, Japan), a multipurpose catheter with curved tip (Slip-Cath Beacon, 100 cm 0.044”, Penumbra, California, United States), and a Penumbra Neuron MAX (0.088", Penumbra) were used in both the simulation (\insilico) and the experimental testbed (\invitro). Only the guidewire was actively manipulated to reach the target, while both catheters remained static in the aortic segments below the arch to provide a stable base that facilitated precise guidewire manipulation.

The experiments were conducted in a mock \gls{or} (\figref{}~\ref{fig:method_experimental_setup}b--d), equipped with a mobile C-arm system (Cios Spin, Siemens Healthineers, Germany) with monoplane fluoroscopic imaging (mode: \textit{vascular standard}, 30 pulses/s). A video stream of the fluoroscopic images was obtained using a video capture card. The tip position and two further points on the guidewire were tracked using a modified tracking algorithm~\cite{2022Eyeberg} and provided to the \agent{} (\secref{}~\ref{subsec_method_agent}). An unaltered video stream was displayed to participants in the adjacent control room, which was separated from the \gls{or} by radiation shielding and equipped with a viewing window into the \gls{or}. All software components required for the experiments—including robot control, hardware communication, tracking, \agent{} inference and response generation, and data logging—were executed on a notebook computer equipped with an Intel® Core™ Ultra 9 185H processor (2.50 GHz), 32 GB RAM, and integrated Intel® Arc™ Graphics, running a 64-bit Windows 11 operating system.

\subsection{Robotic Manipulator and User Control Input Device}\label{subsec_method_robot_inputdevice}
We used a custom-built endovascular robot to independently control the rotation $\varphi_{\mathrm{Rob}}$ and translation $d_{\mathrm{Rob}}$ of one instrument
(\figref{}~\ref{fig:method_experimental_setup}). Building upon a previous design~\cite{sadati2025autonomous}, the robot features an enhanced control architecture with dedicated motor drivers (Tic 36v4, Pololu Corporation, USA).
Participants operated the robot via a device-mimicking \gls{ucid}~\cite{jackson2026} (\figref{}~\ref{fig:method_experimental_setup}b). Through a slider-based mechanism, the interface allows operators to simultaneously input rotation ($\varphi_{\mathrm{Inp}}$) and translation ($d_{\mathrm{Inp}}$) with a single hand movement, thereby facilitating intuitive operation. An additional red push-button functions as a `dead-man switch' (\figref{}~\ref{fig:method_experimental_setup}c), enabling the Full Copilot Assistance functionality (\secref{}~\ref{subsec_method_copilot}).

\subsection{\agent{} for Endovascular Navigation}\label{subsec_method_agent}
The \agent{} was trained exclusively in simulation, using the patient-specific anatomical model and single instrument (i.e., guidewire) control. During training, the agent had no access to the physical robotic system, fluoroscopic imaging and tracking data, or the physical characteristics of the \invitrodash{} testbed (e.g., friction, control delays). 
The learned policy was subsequently deployed directly in the physical environment (\invitro) via zero-shot sim-to-real transfer, without prior hardware-based evaluation or fine-tuning.

Our simulation environment follows previous work based on the stEVE framework and SOFA~\cite{Karstensen2025Learning-basedInterventions,Robertshaw2026TowardHM_MARL}, where a Navien 0.058" support catheter and 0.035" guidewire were modeled using beam-based finite element representations. 
The patient-specific vascular anatomy was generated by fusing neck and body \gls{cta} scans to create complete femoral-to-cerebral arterial trees. 
To improve generalization, vasculatures were augmented through random scaling and rotation, while guidewire–vessel interaction parameters were tuned to replicate realistic device behavior and support subsequent transfer to physical experiments~\cite{Robertshaw2024_auton}. \agent{} observations were intentionally limited to the coordinates of the three most distal tracked guidewire points ($P_1, P_2, P_3 \in \mathbb{R}^2$), the target location, previous device positions, and the previous action, without access to vessel geometry. Using this method, one can later extract these device tip points also from live fluoroscopic images~\cite{Robertshaw2026RealTimeNavigation} during \gls{mt}, providing the inputs needed for the \agent{} to navigate the vasculature in 3D. The action space comprised proximal guidewire rotation and translation commands, constrained to clinically realistic speed limits of 180°/s and 40 mm/s~\cite{Karstensen2025Learning-basedInterventions}, respectively, enabling investigation of autonomous endovascular navigation using only device-tip tracking information.

We employed the \gls{sac} \gls{rl} algorithm~\cite{robertshaw2026safeautonomousroboticendovascular} with multi-worker distributed training to optimize the control policy for the autonomous navigation \agent{}.

The policy and value networks employed recurrent architectures to capture temporal dependencies in the observation space. Both the twin Q-networks and policy network consisted of a \gls{lstm} head followed by a \gls{mlp}, with key architectural and training hyperparameters summarized in Table~\ref{tab:sac_hyperparameters}. Training was performed using \gls{sac} with Adam optimization, linear learning-rate scheduling, and action normalization. A 500K-step warm-up phase using uniformly random actions preceded exploration and policy learning for up to 10M steps, with periodic evaluation following the protocol of~\cite{Karstensen2025Learning-basedInterventions}. 

Success was evaluated using three metrics: task success rate, normalized path ratio, and procedure time~\cite{Robertshaw2026positionstatement}. Task success rate served as the primary criterion for selecting the optimal evaluation checkpoint during \gls{rl} training~\cite{Robertshaw2026TowardHM_MARL}. Based on this metric, the checkpoint corresponding to 8M exploration steps was selected for all subsequent analyses. The \agent{} was then evaluated both \insilico{} and \invitro{} (\gls{rica}: $n = 100$; \gls{reca}: $n = 100$) to assess autonomous performance in the absence of human intervention. Training was conducted on an NVIDIA DGX A100 system (Santa Clara, CA, USA) equipped with eight GPUs and required approximately 11 hours to complete 8M exploration steps.

\begin{table}[t] \centering \caption{Key hyperparameters for soft actor-critic algorithm.} \label{tab:sac_hyperparameters} \begin{tabular}{l c l c} \hline \textbf{Parameter} & \textbf{Value} & \textbf{Parameter} & \textbf{Value} \\ \hline LSTM size & 500 & Batch size & 32 \\ MLP layers & 4 & Replay buffer & $10^6$ \\ MLP hidden units & 900 & Exploration noise & 0.25 \\ Learning rate & $2.2\times10^{-4}$ & Training steps & 10M \\ Discount factor $\gamma$ & 0.99 & Warm-up steps & 500K \\ Target update rate $\tau$ & 0.005 & Update ratio & 0.05 \\ 
\hline 
\end{tabular}

\vspace{1mm}
\footnotesize{LSTM, long short-term memory; MLP, multi-layer perceptron.}
\end{table}

\subsection{\CamsWrittenOut}\label{subsec_method_copilot}
AI-driven robotic navigation can be understood as existing along a continuum of autonomy, ranging from manual control of robotic instruments, through shared or semi-autonomous assistance, and ultimately toward fully autonomous navigation. Along this spectrum, the role of the clinician gradually shifts from low-level control of endovascular devices toward higher-level supervision and decision-making, while the AI system assumes increasing responsibility for motion generation and trajectory execution. 

\begin{figure}
    \centering
    \includegraphics[width=1\linewidth]{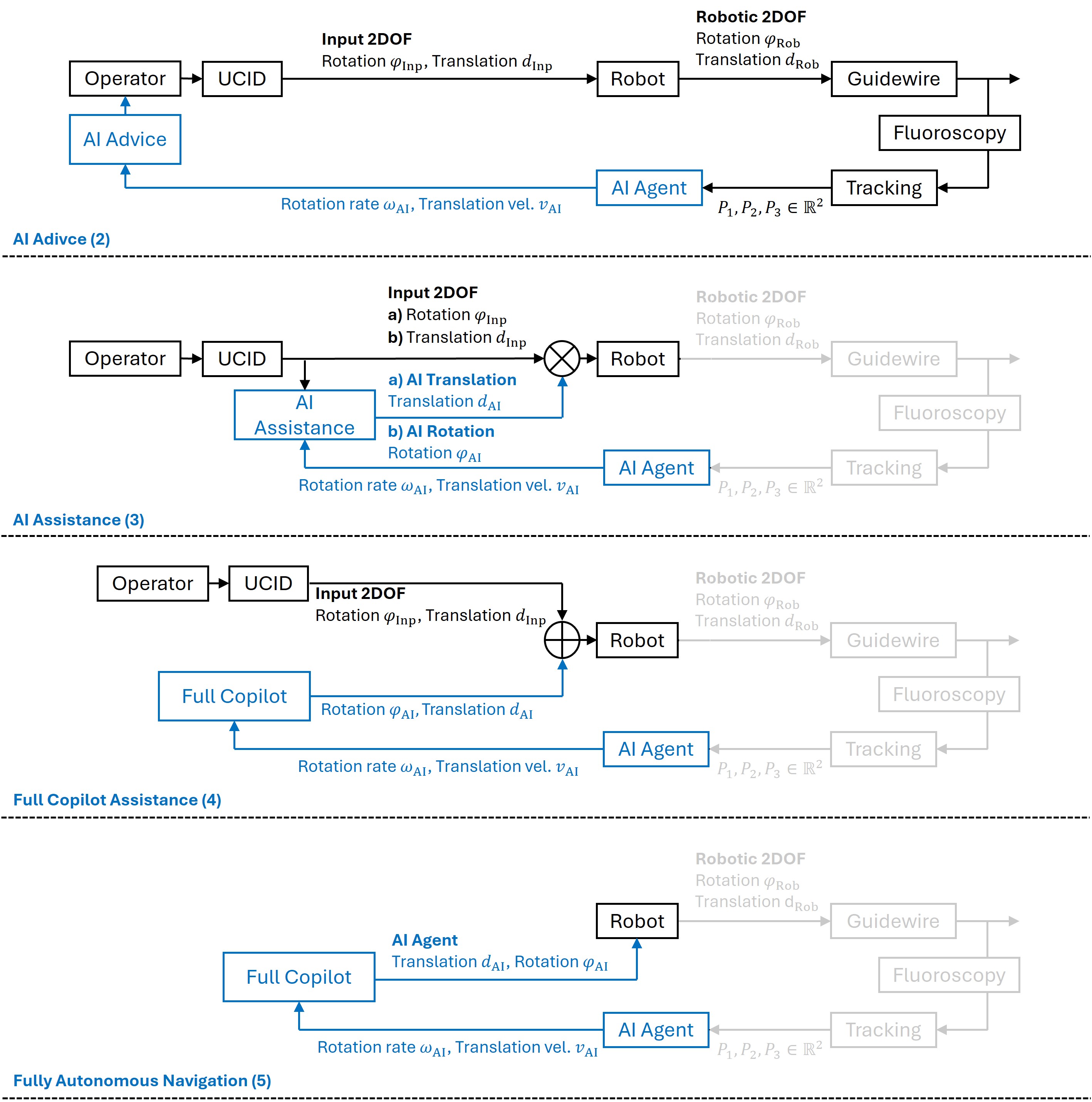}
    \caption{\textbf{Control Architectures for \AIDrivenModes:}
  Collaborative control loops for \CamsWrittenOut{} (2)--(5), with AI components highlighted in blue. The operator controls the robot via the User Controller Input Device (UCID), which manipulates the guidewire within the 3D-printed vascular phantom. Instrument tip coordinates are extracted from the fluoroscopic image stream and provided to the \agent{}. Grey components in \glspl{cam} (3)--(5) denote identical components. The symbol ``$\otimes$'' represents the combination of inputs in \camthree{}, whereas ``$\oplus$'' denotes a mutually exclusive selection between inputs in \camfour{}.
    }
    \label{fig:method_control_architectures}
\end{figure}

To systematically investigate how control should be shared between clinicians and AI systems, we introduced \glspl{cam}, a framework for studying clinician-AI collaboration across progressively increasing levels of autonomy (\figref{}~\ref{fig:introduction_visual_abstract_cam_modes}). These interaction modes were informed by discussions with our clinical partners and represent clinically motivated approaches to sharing control between human operators and AI systems.
Each mode corresponded to a specific allocation of control authority between the operator and the \agent{}, thereby enabling the isolated evaluation of guidance, shared control, delegated control, and fully autonomous navigation paradigms (\tableref~\ref{tab:copilot_assistance_modalitites}, \figref{}~\ref{fig:method_control_architectures}).
\glspl{cam} (0) and (1) constituted conventional modes while \glspl{cam} (2)--(5) were \AIdrivenmodes. 

\subsubsection*{\textbf{\camnull}} As baseline, the operator controlled the guidewire hands-on, i.e. manually $[\varphi_{\mathrm{Man}}, d_{\mathrm{Man}}]$ from within the \gls{or}.
\subsubsection*{\textbf{\camone}} As robotic baseline, the operator controlled the robot (i.e., teleoperation) with the \gls{ucid} providing $\varphi_{\mathrm{Inp}}$ and $d_{\mathrm{Inp}}$ from a radiation-protected control room (\figref{}~\ref{fig:method_experimental_setup}b). 
\subsubsection*{\textbf{\camtwo}}
\begin{figure}
    \centering
    \includegraphics[width=1\linewidth]{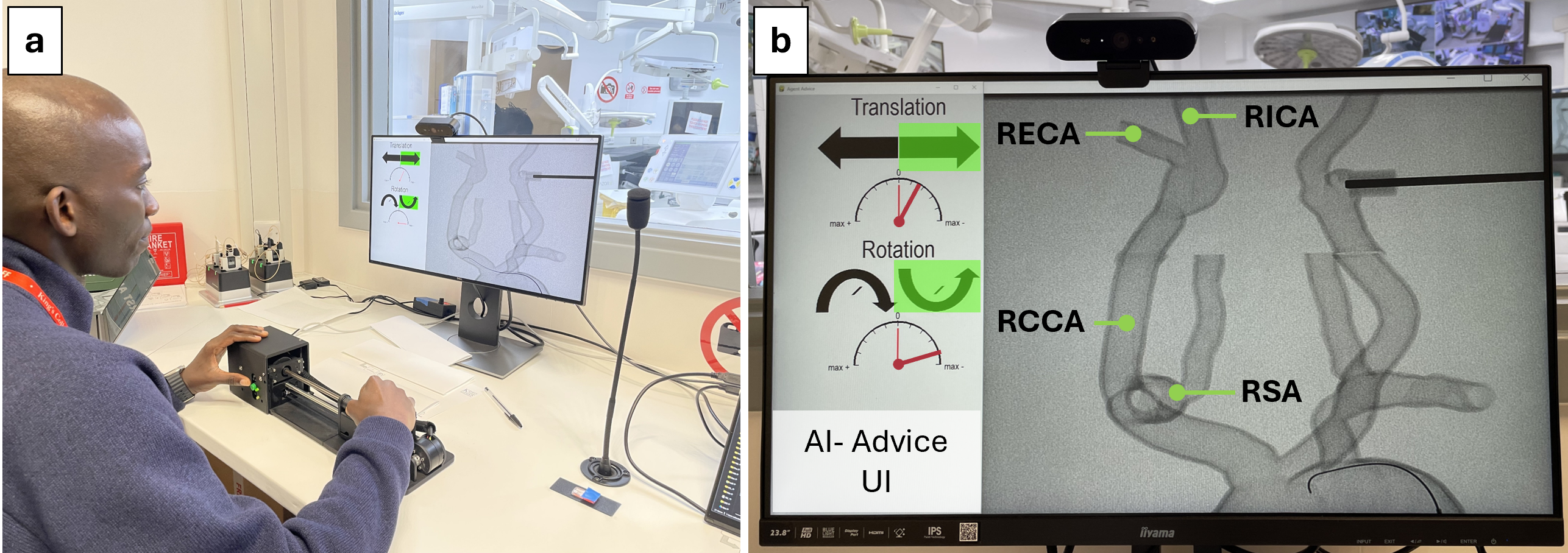}
    \caption{
    \textbf{AI-Advice User Interface (UI)}: (a) Participant navigating the guidewire via robotic manipulator with the User Controller Interface Device (UCID) from the control room. 
    (b) AI Advice is provided by the \gls{ui} next to the fluoroscopic image. Direction of motion is highlighted in green for both translation and rotation while the red needle in the dial gauges quantify the amplitude of motion. 
    Note, that labels for the right subclavian artery (RSA), right common carotid artery (RCCA), right external carotid artery (RECA), and right internal carotid artery (RICA) were not shown on the \gls{ui} to the operator during the experiments.
    }
    \label{fig:method_ai_advice_panel}
\end{figure}

This mode represented perceptual assistance. The operator retained full control of both DoF while the \agent{} only provided action recommendation via the imaging video stream enhanced with an \gls{ai}-Advice \gls{ui} (\figref{}~\ref{fig:method_ai_advice_panel}a--b). 
\gls{ai}-Advice indicated recommended direction and normalized amplitude of movement, based on the recommended normalized actions for rotation rate  $\omega_{\mathrm{AI}}$ and translation velocity $v_{\mathrm{AI}}$ of the \agent{}.
\subsubsection*{\textbf{\camthree}} 
This mode represented shared control through DoF allocation.
The \agent{} controlled one robotic DoF, i.e. either \camthreea{} or \camthreeb. The remaining DoF was controlled by the operator. The operator's position commands $[\varphi_{\mathrm{Inp}}$, $d_{\mathrm{Inp}}]$ served as the primary input, whereas the velocity actions generated by the \agent{} $[\omega_{\mathrm{AI}}$, $v_{\mathrm{AI}}]$ were projected onto, and combined with, the operator's position command, resulting in a combined control input $[\varphi_{\mathrm{Inp}}, d_{\mathrm{AI}}]$ for \camthreea{} and $[\varphi_{\mathrm{AI}}$, $d_{\mathrm{Inp}}]$ for \camthreeb.

\subsubsection*{\textbf{\camfour}} 
This mode represented temporary task delegation. The \agent{} assumed control of both robotic DoF with $[\varphi_{\mathrm{AI}}, d_{\mathrm{AI}}]$ whenever activated by the operator through holding the \copilotPushButton{} on the \gls{ucid} (\figref{}~\ref{fig:method_experimental_setup}c).
In \camfoura{}, full assistance was activated at a predefined point, spanning the segment from the aortic arch to the \gls{rcca}. In \camfourb{}, the operator could freely choose when to activate and deactivate full assistance.

\subsubsection*{\textbf{\camfive}}
This mode represented complete task delegation. The \agent{} navigated autonomously from start to target without human interaction.

Together, the \glspl{cam} provided a structured framework for evaluating how different distributions of control authority influence objective performance and subjective operator perception, thereby enabling systematic investigation of clinician-AI collaboration in robotic neuroendovascular intervention.

\begin{table}[!t]
\caption{Taxonomy of \CamsWrittenOut{}}
\label{tab:copilot_assistance_modalitites}
\centering
\scriptsize
\setlength{\tabcolsep}{3pt}
\renewcommand{\arraystretch}{1.05}
\begin{tabular}{llll}
\hline
 \# & CAM & Rotation & Translation  \\
\hline
0 & Manual (baseline) & Manual $\varphi_{\mathrm{Man}}$ & Manual $d_{\mathrm{Man}}$ \\
1 & Direct Robot Control & Teleoperated $\varphi_{\mathrm{Inp}}$  & Teleoperated $d_{\mathrm{Inp}}$    \\
2 & AI Advice & Teleoperated $\varphi_{\mathrm{Inp}}$  & Teleoperated $d_{\mathrm{Inp}}$    \\
\hline
\textbf{3} & \textbf{AI Assistance}& \multicolumn{2}{c}{{Splitting Robotic DOF}}\\
3a & AI Translation & Teleoperated $\varphi_{\mathrm{Inp}}$ & \agent{}  $d_{\mathrm{AI}}$   \\
3b & AI Rotation & \agent{} $\varphi_{\mathrm{AI}}$ & Teleoperated $d_{\mathrm{Inp}}$  \\
\hline
\textbf{4} &\textbf{Full Copilot Assistance} & \multicolumn{2}{c}{{Splitting Navigation Path}}\\
4a & Predefined Sections & $[\varphi_{\mathrm{Inp}}, \varphi_{\mathrm{AI}}]$& $[d_{\mathrm{Inp}}, d_{\mathrm{AI}}]$\\
4b & Operator-Defined Sections & $[\varphi_{\mathrm{Inp}}, \varphi_{\mathrm{AI}}]$& $[d_{\mathrm{Inp}}, d_{\mathrm{AI}}]$\\
\hline
5 & Fully Autonomous Navigation& \agent{} $\varphi_{\mathrm{AI}}$  & \agent{} $d_{\mathrm{AI}}$ \\
\hline
\end{tabular}
\end{table}

\subsection{Hypothesis and Statistical Analysis}\label{subsec_method_hypothesis_statistics}
We hypothesized that shared autonomy and copilot-assisted control modes improve navigation performance relative teleoperated control while maintaining the clinician's sense of agency and supervisory control. Accordingly, we assessed whether \camfour{} enhances navigation performance relative to \camone, using \camnull{} as the reference baseline condition.
Informed by the recommendations of the Stakeholder Taskforce for AI-assisted Robotic Thrombectomy (START) position statement~\cite{Robertshaw2026positionstatement}, we evaluated navigation performance using complementary measures of task success, navigation efficiency, and navigation errors.
The primary outcome metrics comprised task success rate (i.e., reaching the target anatomy) and \gls{t2t}, which served as the measure of navigation effectiveness and efficiency.
The secondary outcome metric was the number of navigation errors, reflecting deviations from the intended navigation path during vessel selection and target approach. Navigation errors were defined as any instance of probing or navigating into an incorrect vessel branch. Specifically, errors comprised i) overshooting into the ascending aorta towards the aortic valve, ii) navigating into the \gls{rsa}, and iii) falsely probing \gls{rica} or \gls{reca}.

To ensure sufficient statistical power for detecting differences in \gls{t2t}, the required number of repetitions per condition was determined through an \textit{a priori} power analysis (\supmaterial{}~A). 
The study was planned as a within-subject design with $N=10$ participants and 6 repetitions per condition, allowing sufficient power to detect differences in navigation time and error count between \camfour{} and the conditions \camnull{} and \camone.

The remaining \glspl{cam}, \camtwo{} and \camthree, were included as exploratory conditions with 4 repetitions per participant.

\gls{t2t} and errors were analyzed using successful trials only. Due to the expected small number of participants, differences among \glspl{cam} were assessed using a permutation-based one-way ANOVA. To account for the study design, \gls{cam} condition assignments were permuted 10,000 times while preserving the participant-cohort structure. The ANOVA F statistic was used to evaluate whether the mean outcome differed across \glspl{cam}. Statistical significance was assessed at $\alpha=0.05$. Post-hoc pairwise comparisons were conducted using two-sided permutation tests, and the Holm step-down correction was applied to control the family-wise error rate across comparisons within each outcome. Corrected values are reported as $p_{\mathrm{Holm}}$, with statistical significance defined as $p_{\mathrm{Holm}}<0.05$.

\subsection{Participants}\label{subsec_method_participants}
The final participant cohort ($N=13$) consisted of 10 neurointerventional radiologists, including 6 UK consultants (US attending equivalents) and 4 fellows with 2--18 years of neurointerventional experience, 1 general interventional radiology resident, and 2 third-year medical students. Participants were stratified into four subgroups based on procedural target assignment and experience level. The participant briefing is provided in \supmaterial{}~B.

\subsection{Experimental Procedure}\label{subsec_method_experimental_procedure}
The objective of the task was to navigate the guidewire from its starting point in the aortic arch into the target anatomy (\gls{rica} or \gls{reca}) under different \glspl{cam}. A maximum time limit of \SI{240}{\second} was imposed for each experimental run.
To mitigate potential ordering and learning effects, the sequence of evaluated \glspl{cam} was randomized across participants, with the exception of \camnull, which was consistently performed last.

\subsection{Survey Design}\label{subsec_method_survey_design}
Following the experimental trials, participants completed a survey designed to capture subjective evaluations of the different \glspl{cam}. The survey assessed perceived workload, usability, and overall preference for each in order to complement the objective performance measures collected during the experiments.
Subjective workload was assessed using a modified version of the \gls{tlx}. To reduce respondent burden and facilitate repeated assessments within the surgical simulation workflow, all workload dimensions were rated on a 5-point Likert scale (1 = low workload, 5 = high workload).

Usability was assessed using the \gls{sus}, a standardized and widely adopted 10-item questionnaire with 5-point Likert-scale responses~\cite{brooke1996sus}. The \gls{sus} provides a reliable measure of perceived usability and ease of interaction with each \gls{cam} and responses were converted to the standard \gls{sus} score ranging from $0$--$100$, with higher scores indicating better perceived usability. Usability outcomes were interpreted according to the ISO 9241-11 definition of usability, encompassing effectiveness, efficiency, and user satisfaction in the specified context of use \cite{iso9241}. 

Statistical analysis of \gls{tlx} and \gls{sus} mirrored that of \gls{t2t} described in \secref{}~\ref{subsec_method_hypothesis_statistics}, comprising a two-sided permutation-based one-way ANOVA followed by pairwise two-sided permutation tests with Holm step-down correction.

In addition to the standardized questionnaires, participants were asked to comparatively rank the evaluated \gls{cam} according to their overall preference and perceived effectiveness. This ranking was included to capture qualitative operator preferences that may not be fully reflected by workload or usability scores alone.

Additional survey items assessed participants’ experiences with and perceptions of the \gls{ucid}, as well as their evaluation of the realism of the \invitrodash{} phantom. 

\section{Results} \label{sec_results}
\subsection{Objective Performance Metrics}\label{subsec_results_objective_metrics}
\subsubsection{\textbf{Success Rate}}\label{subsubsec_results_successrate}
\begin{figure}
    \centering
    \includegraphics[width=1\linewidth]{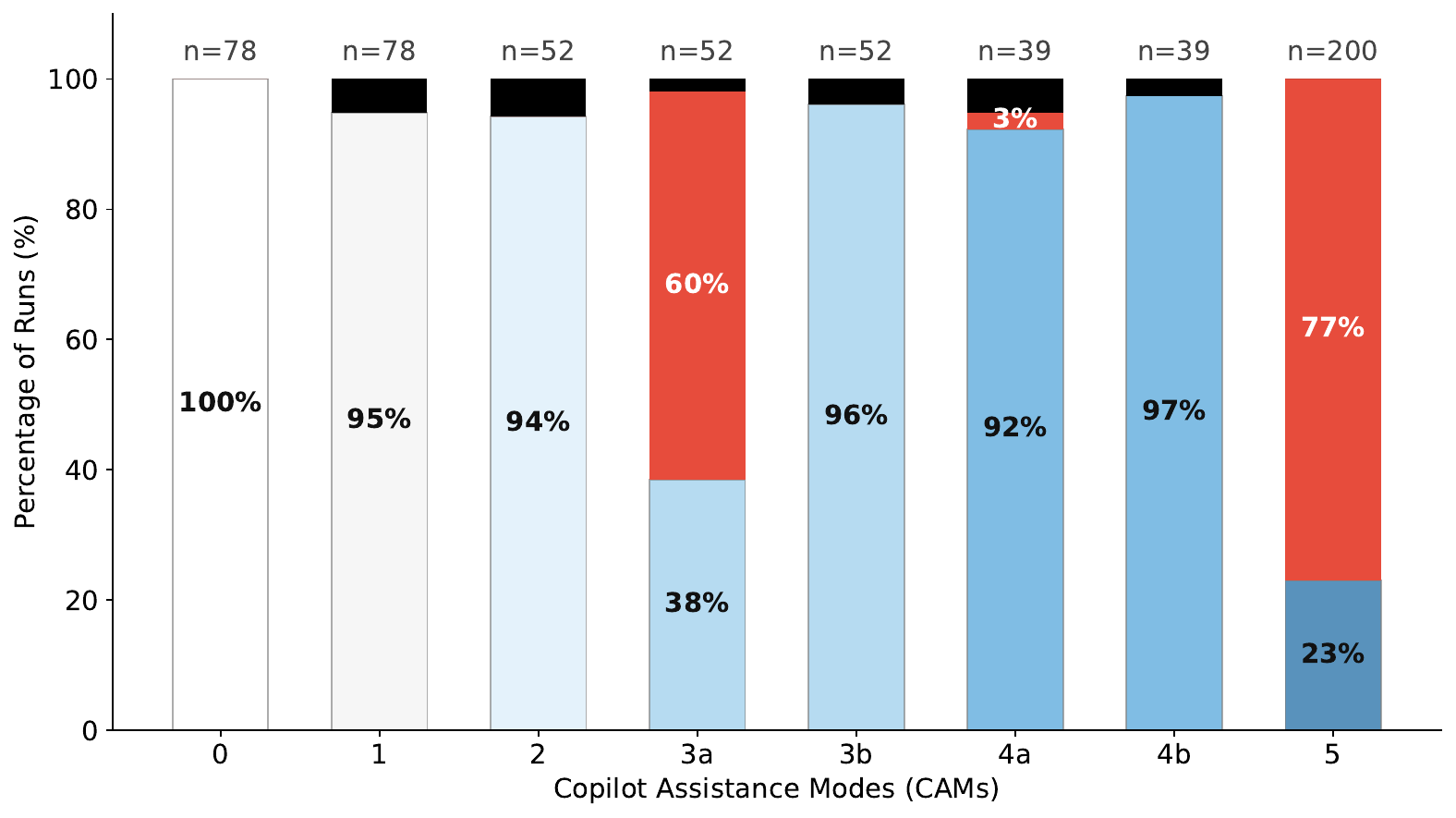}
    \caption{\textbf{Success Rates Comparison:} 
    Stacked bars show the proportions of successful, failed (red), and timed-out (black) runs for each of the \CamsWrittenOut{} (\tableref~\ref{tab:copilot_assistance_modalitites}). Percentages within the bars indicate success and failure rates, and values above the bars indicate the total number of runs (n). For each participant, \glspl{cam} (0), (1), and (4), i.e. 4a and 4b combined, were tested in 6 runs, whereas \glspl{cam} (2), (3a), and (3b) were each tested in 4 runs. \camfive{} was evaluated without participant involvement in 200 autonomous runs (100 targeting the right internal carotid artery (RICA) and 100 targeting the right external carotid artery (RECA)). \colorgradientWhiteGreyBlue{}
    }
    \label{fig:results_successrate}
\end{figure}

\figref{}~\ref{fig:results_successrate} shows the run outcomes for all \glspl{cam}. 
\camfive{} achieved the lowest success rate (23\%), followed by \camthreea{} (38\%). All remaining robotic \glspl{cam} achieved success rates between 92\% (\camfoura) and 97\% (\camfourb{}) while \camone{} without any AI assistance gave 95\%.
All robotic \glspl{cam} exhibited time-out rates between 2\%--6\% including \camone{}. Compared with \camfive, all assisted and manual conditions demonstrated substantially improved performance. Fails in \camthreea{} occurred when the \agent{} trapped the instrument in an incorrect vessel branch. A single fail run also occurred in \camfoura{}, when the \agent{} autonomously retracted the instrument from the scene. 
An in-depth analysis for \camfive{} can be found in the \supmaterial{}~C.

\subsubsection{\textbf{Time-to-Target (T2T)}}\label{subsubsec_results_t2t}
\begin{figure}
    \centering
    \includegraphics[width=1\linewidth]{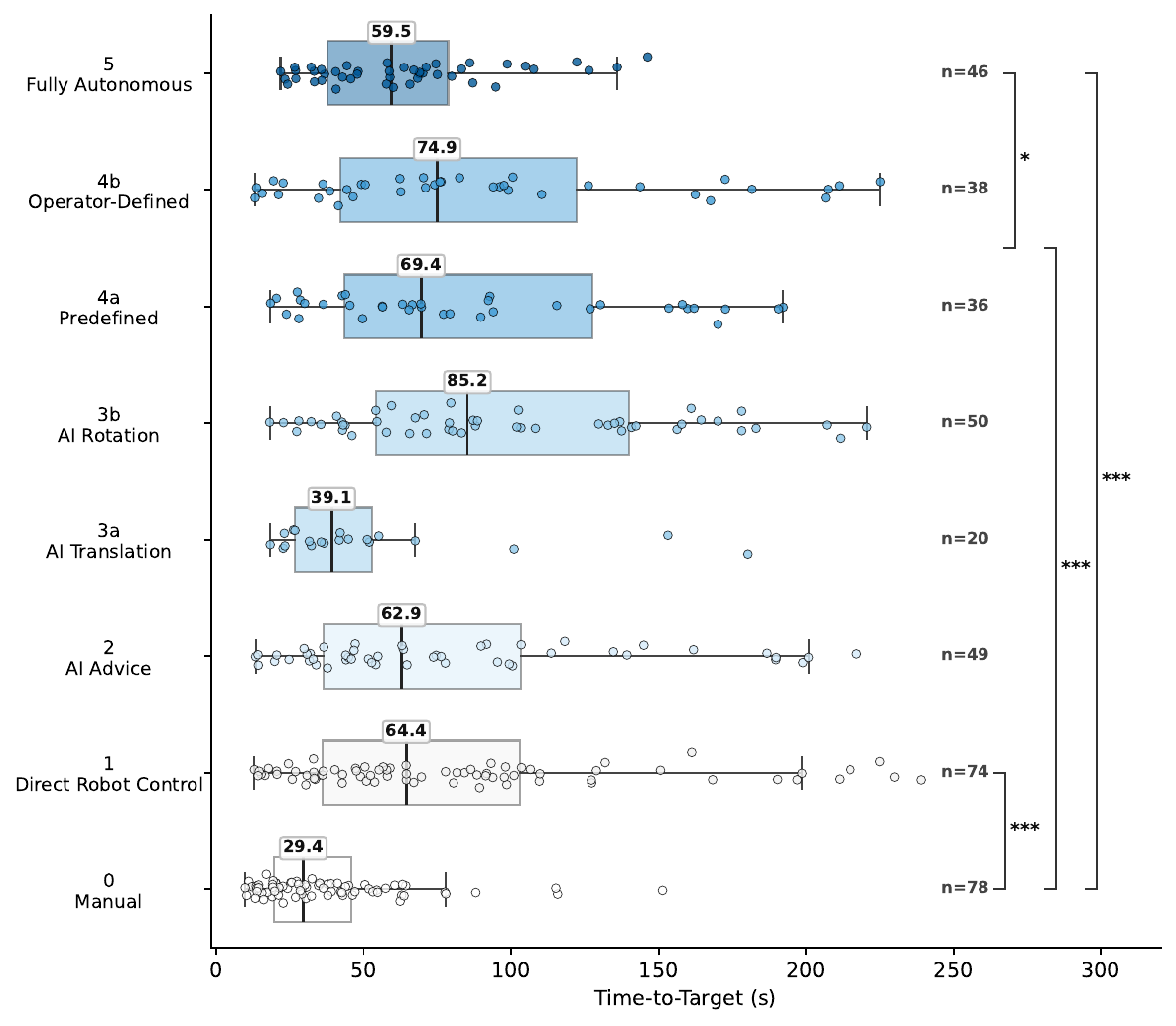}
 \caption{\textbf{Time-to-Target (T2T):}
 Distribution of T2T for all successful runs of each Copilot Assistance Mode (CAM).
 n values denote the number of successful runs per condition. Significance levels are denoted based on Holm-adjusted p-values: ($\ast$), ($0.01\leq p_{\mathrm{Holm}}<0.05$); ($\ast\ast\ast$), ($p_{\mathrm{Holm}}<0.001$).
\colorgradientWhiteGreyBlue{}
}
    \label{fig:results_t2t}
\end{figure}

\figref{}~\ref{fig:results_t2t} shows the distribution of \gls{t2t}. Overall, there was no significant difference for \gls{t2t} between \camone{} and any of the AI modes (2)--(4) that were not fully autonomous. 
Medians ranged from \SI{29.5}{\second} for \camnull{} to \SI{85.2}{\second} for \camthreeb.
A permutation-based ANOVA revealed a significant effect of \gls{cam} on \gls{t2t}, $F(6,338) = 10.643$, $p_{\mathrm{perm}} < 0.0001$ with a large omnibus effect, $\eta^2=0.159$ and $\omega^2=0.144$. Holm-corrected pairwise comparisons identified two set of significant effects.
First, \camnull{} resulted in significantly lower \gls{t2t} values than \camone{}, \camfour{}, and \camfive{} (all $p_{\mathrm{Holm}}=0.0006$). The corresponding median \gls{t2t} values were \SI{29.45}{\second}, \SI{64.40}{\second}, \SI{72.15}{\second}, and \SI{59.50}{\second}, respectively. Second, a significant difference was observed between medians of \camfour{} and \camfive{} (\SI{72.15}{\second} vs.~\SI{59.50}{\second}; $p_{\mathrm{Holm}}=0.0303$). 
However, the \gls{t2t} of \camthreea{} and \camfive{} should be interpreted cautiously, given the relatively small number of successful runs. 

\begin{figure*}[t]
    \centering
    \includegraphics[width=\textwidth]{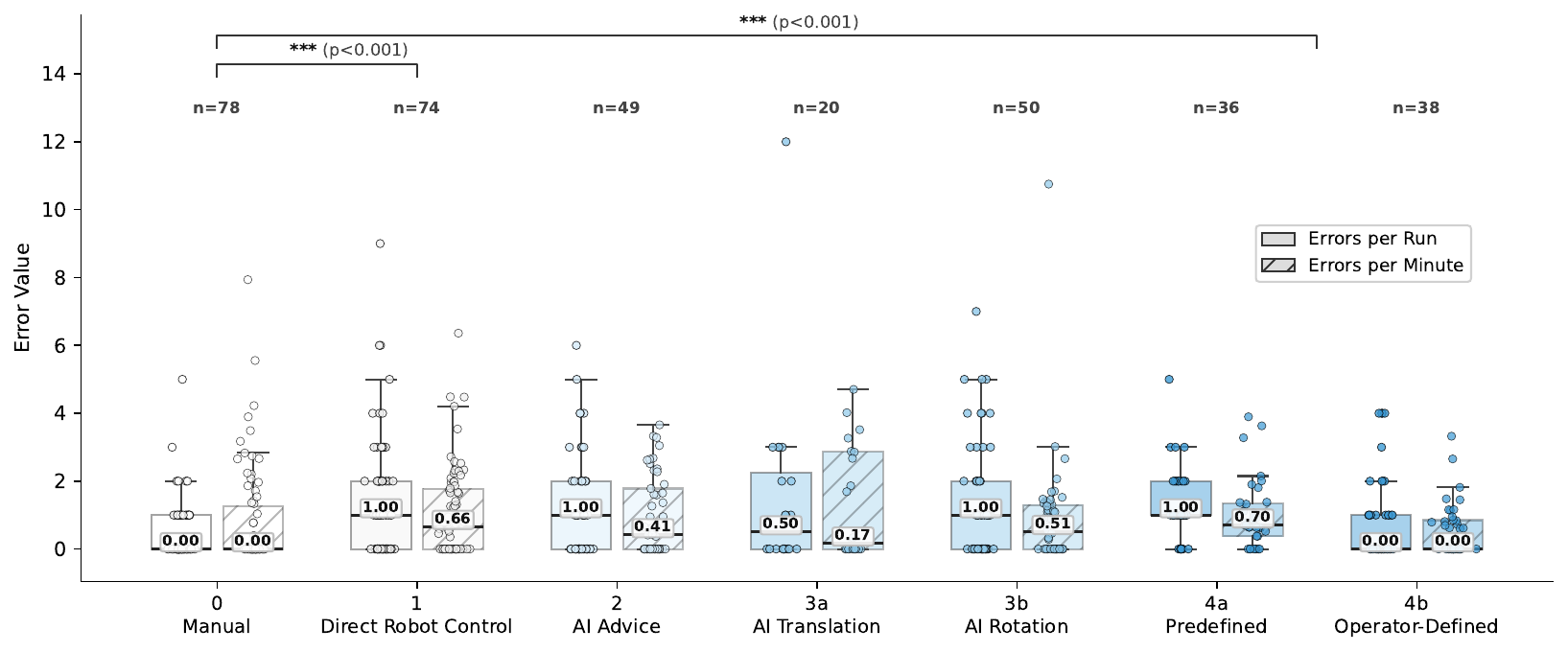}
    \caption{\textbf{Errors:}
    Distribution of navigation errors per run (filled) and per minute (hatched) for all successful runs of each Copilot Assistance Mode (CAM). Navigation errors correspond to deviations into incorrect vessel branches (\secref{}~\ref{subsec_method_hypothesis_statistics}). 
    n denotes the number of successful runs per \gls{cam}. \camfive{} is absent because any navigation error resulted in an unsuccessful trial. Unlike the clinician-supervised modes, in this mode, the \agent{} could not recover from these errors, leaving no successful trials to display. \colorgradientWhiteGreyBlue{}
    }
    \label{fig:result_errors}
\end{figure*}

\subsubsection{\textbf{Errors}}\label{subsubsection_results_errors}
\figref{}~\ref{fig:result_errors} shows the distribution of navigation errors per run and per minute. Navigation errors, as defined in \secref{}~\ref{subsec_method_hypothesis_statistics}, comprise deviations into incorrect vessel branches or bifurcations. 
All pairwise comparisons between \camone{} and the the \AIdrivenmodes{} (2), (3a), (3b), (4a), and (4b) were non-significant ($p = 0.590$, $0.761$, $0.836$, $0.802$, and $0.191$, respectively).

Median error counts ranged from $0.00$--$1.00$ errors per run, while median error rates ranged from $0.00$--\SI{0.70}{\per\minute}. The lowest values were observed for \camnull{} and \camfive, whereas the highest values occurred in \camfoura.
For \glspl{cam} (0)--(5), a permutation-based ANOVA revealed a significant effect of \gls{cam} on errors per run ($F(6,338)=3.818$, $p_{\mathrm{perm}}=0.0008$, $\eta^2=0.063$, $\omega^2=0.047$), but not on errors per minute ($F(6,338)=1.179$, $p_{\mathrm{perm}}=0.3148$, $\eta^2=0.021$, $\omega^2=0.003$). 
Holm-corrected pairwise tests for errors per run showed that \camnull{} differed from \camone{} and \camfour{} (both $p_{\mathrm{Holm}}=0.0003$). 
Median errors per run were $0.00$ for \camnull{} and $1.00$ for both \camone{} and \camfour{}. No significant difference was found between \camone{} and \camfour{} ($p_{\mathrm{Holm}}=0.2965$). Likewise, no significant differences were observed among \camtwo, \camthreea, and \camthreeb, whose median errors per run were $1.00$, $0.50$, and $1.00$, respectively (corresponding median errors per minute: \SI{0.41}{\per\minute}, \SI{0.17}{\per\minute}, and \SI{0.51}{\per\minute}).

\subsubsection{\textbf{AI Usage Analysis}}\label{subsubsec_results_ai_usage_analysis}
\begin{figure}[t]
    \centering
    \includegraphics[width=1\linewidth]{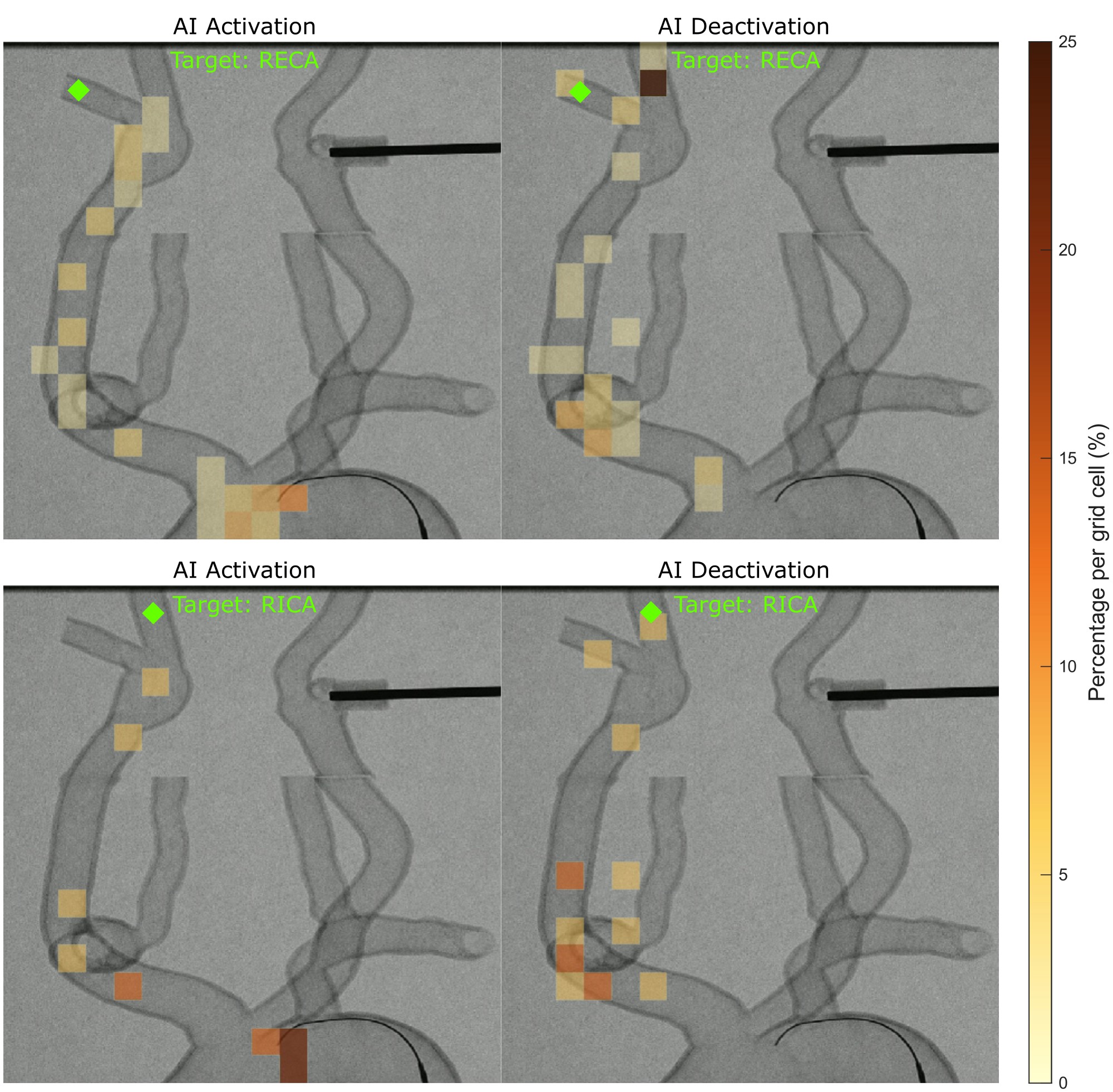}
    \caption{\textbf{AI activation and deactivation during \camfourb{}:} Heatmaps depict the spatial distribution of AI activations (left column) and deactivations (right column) during navigation to the right external carotid artery (RECA, top row) and right internal carotid artery (RICA, bottom row). Values represent the percentage of AI-assisted interactions per grid cell and are overlaid on fluoroscopic phantom images. Warmer colors indicate a higher frequency of AI activation or deactivation within the corresponding region.
    }
    \label{fig:results_ai_usage_heatmap}
\end{figure}

For \camfourb, where participants could freely use the \agent{} throughout the task, we additionally analyzed the spatial distribution of \agent{} activity by recording the start and end positions of each utilization interval. This allowed us to characterize participants' engagement with the \agent{} and quantify its contribution during the navigation task.
\figref{}~\ref{fig:results_ai_usage_heatmap} shows the spatial distribution of AI activation and deactivation. AI was most frequently activated at the start of the procedure in the aortic arch, and to a lesser extent around the \gls{rsa}-\gls{rcca} bifurcation.
Deactivation points formed a dense cluster in the \gls{rica} (for \gls{reca} targets) and another near the \gls{rsa}-\gls{rcca} bifurcation including within the \gls{rsa} origin.

\subsection{Survey}\label{subsec_results_survey}
To ensure that the subjective evaluations reflected the views of individuals with relevant clinical expertise, only those participants with an interventional radiology background were included (11/13).
The complete survey results, including all free-text responses, are available in the \supmaterial{}~D.
\subsubsection{Robot Controller Interface 
\gls{ucid}}\label{subsubsec_results_ucid_rating}
Several clinicians (64\%, $7/11$) reported prior exposure to the \gls{ucid}~\cite{jackson2026}, while $4/11$ reported first-time use. 
Free-text responses indicated that the interface was intuitive, easy to learn, and reasonably aligned with clinically familiar guidewire manipulation with continuous control. It was also valued for the ability to coordinate translation and rotation with one hand, highlighting its potential for practical application, e.g., performing additional tasks with the other hand. 
In contrast, participants reported limitations related to delays between input commands and robot movement, particularly during rotational instrument manipulation; the restricted range of continuous translational motion imposed by the \gls{ucid} stroke length; and reduced tactile feedback.

\subsubsection{Phantom Realism}\label{subsubsect_results_phantom_realism}
The phantom was rated as realistic by most clinicians, with a mean realism score of $4.1/5.0$, median score of $4.0/5.0$ and range [$3.0$--$5.0$]. Most clinicians (82\%, $9/11$) rated realism as $4.0$ or $5.0$. According to free-text responses, the positive ratings were mainly due to the vessel geometry, branch localization, overall anatomical shape, and guidewire behavior. Functionally, it was described as similar to real neurointerventional navigation, including familiar challenges such as the guidewire preferentially entering the \gls{rsa}. However, participants also identified important limitations of the phantom model as not fully reproducing \invivodash~conditions~\cite{Robertshaw2026positionstatement}, particularly blood flow, pulsatility, vessel compliance, smaller branch detail, and realistic friction. 

\subsubsection{NASA-TLX}\label{subsubsec_results_survey_nasatlx}
\begin{figure}
    \centering
    \includegraphics[width=1\linewidth]{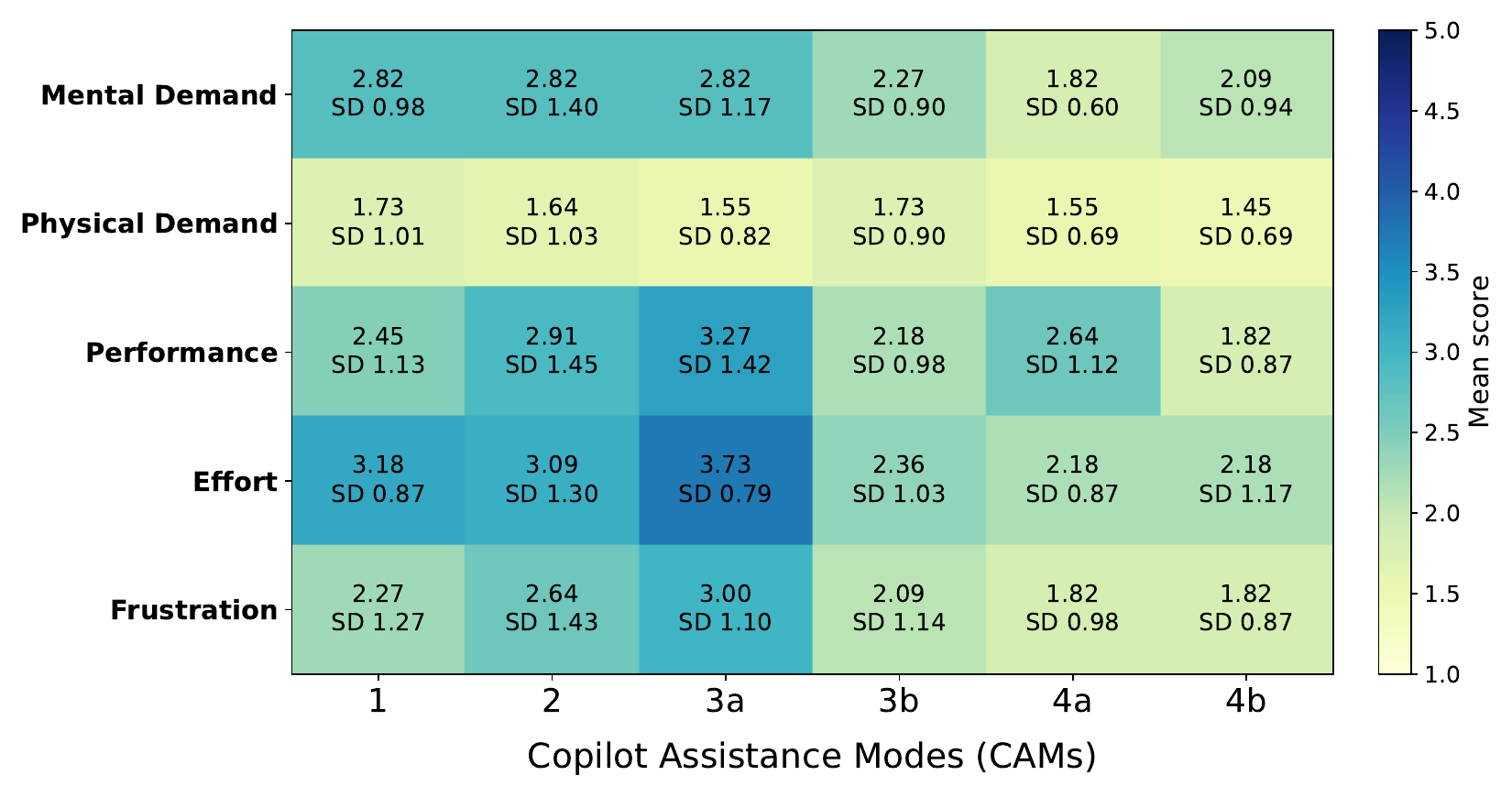}
    \caption{\textbf{Subjective Workload:}
    Heatmap of NASA-TLX ratings by subscale and condition. Values in each cell are mean ± SD (displayed on two lines) on a 1--5 scale, with the Performance subscale reverse-coded to align interpretation (higher score = higher perceived workload).
    }
    \label{fig:results_nasatlx_heatmap}
\end{figure}

The heatmap in \figref{}~\ref{fig:results_nasatlx_heatmap} illustrates the perceived workload across control conditions. Physical demand remained consistently low across all modes, with mean scores ranging from $1.45$--$1.73$ on a 5-point scale, indicating that neither robotic control nor AI assistance imposed substantial physical burden during guidewire manipulation. Greater variation was observed for mental demand, effort, frustration, and perceived performance. A permutation-based one-way ANOVA revealed an overall effect of control condition on NASA-TLX scores ($F(5,60)=2.77$, $p_{\mathrm{perm}}=0.0217$), with a moderate effect size ($\eta^2=0.187$; $\omega^2=0.118$). However, no pairwise comparison remained statistically significant after Holm correction for multiple comparisons (all $p_{\mathrm{Holm}}\geq0.05$). The smallest adjusted p-value was observed for the comparison between \camthreeb{} and \camtwo{} ($p_{\mathrm{Holm}}=0.0825$). Therefore, the differences described below should be interpreted as descriptive trends rather than statistically significant pairwise effects.
Among the \AIdrivenmodes, \camthreea{} exhibited the highest mean workload ratings, including effort ($3.73 \pm 0.79$), frustration ($3.00 \pm 1.10$), and reverse-coded performance workload ($3.27 \pm 1.42$). This trend is consistent with qualitative feedback suggesting that operators found it challenging to coordinate manual rotational input with autonomous translational motion. In contrast, \camthreeb{} showed a trend toward lower ratings of mental demand, effort, frustration, and performance-related workload -- especially when compared to \camone{}. This trend is consistent with qualitative feedback suggesting that the operator could wait for the AI to rotate correctly before the operator translated the guidewire forwards into the correct bifurcation branch.
The lowest overall workload ratings were observed for \camfour, with both \camfoura{} and \camfourb{} associated with low levels of mental demand, effort, frustration, and performance-related workload.

\subsubsection{System Usability Scale (SUS)}\label{subsubsec_results_survey_sus}
\begin{figure}
    \centering
    \includegraphics[width=1\linewidth]{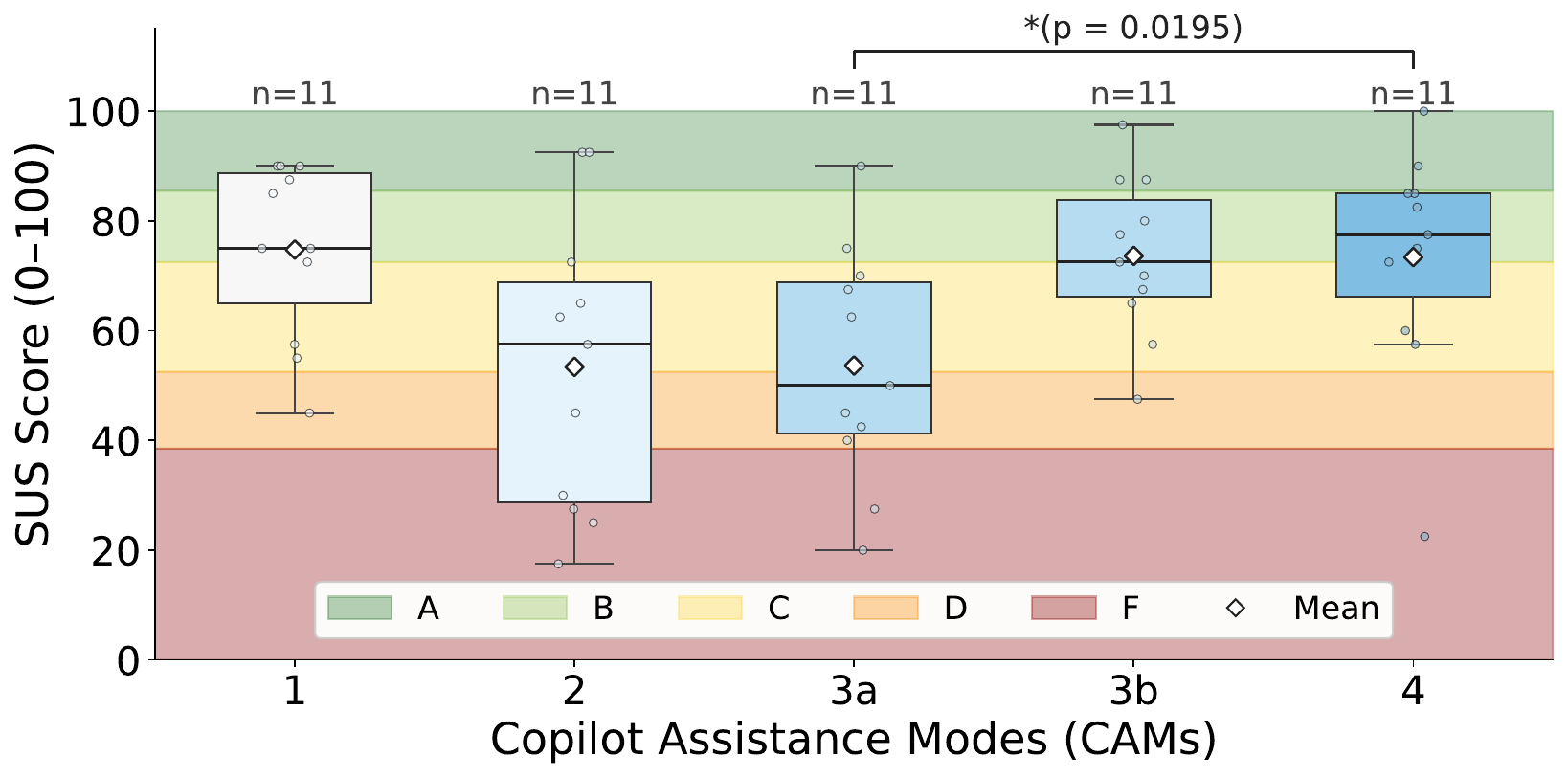}
\caption{\textbf{System Usability:} 
Distribution of System Usability Score (SUS) for all \gls{cam} (expert clinician participants only, $N=11$): boxplots show mean, median, and \gls{iqr} (whiskers: 1.5×IQR); points show individual participants, shaded bands indicate standard \gls{sus} grade ranges (A--F). The \gls{sus} ratings for each item ranged from $0$--$5$ and were converted to a $0$--$100$ scale. \colorgradientGreyBlue{}
}
    \label{fig:SUS}
\end{figure}

\figref{}~\ref{fig:SUS} and Table~\ref{tab:SUS} summarize the \gls{sus} scores across modes. 
Scores were converted to a $0$--$100$ scale and interpreted using standard \gls{sus} benchmarks, where mean values above $68$ indicate above-average usability and values between $50$--$68$ indicate marginal usability. Manual operation was not assessed using the \gls{sus}. Mean \gls{sus} scores ranged from $53.4 \pm 26.5$ (\camtwo{}) to $74.8 \pm 15.9$ (\camone{}). According to \gls{sus} benchmark classifications, \camone{}, \camthreeb{}, and \camfour{} achieved mean scores in the good usability range, whereas \camtwo{} and \camthreea{} fell within the marginal usability range. Across 11 participants and five \glspl{cam}, a permutation-based one-way repeated-measures ANOVA revealed a significant effect of setup on \gls{sus} scores ($F(4,40)=4.17$, $p_{\mathrm{perm}}=0.0076$, $\eta^2=0.29$, $\omega^2=0.22$). Holm-corrected comparisons identified a significant difference only between \camthreea{} and \camfour{} ($p_{\mathrm{Holm}}=0.021$). 

When considering median \gls{sus} scores, \camfour{} achieved the highest overall usability rating (77.5), followed by \camone{} (75.0) and \camthreeb{} (72.5), consistent with the ranked order shown in Table~\ref{tab:SUS}.

Qualitative feedback was broadly consistent with these findings. 
Users reported that \camthreeb{} and \camfour{} reduced workload and improved navigation during challenging procedural segments, although concerns regarding transparency, occasional unexpected AI behavior, and overall system reliability remained.
Participants rated \camone{} favorably but noted the loss of tactile feedback and limited advantages over manual control for experienced operators. The comparatively lower \gls{sus} scores of \camtwo{} and \camthreea{} were frequently attributed to difficulties integrating AI advice into real-time navigation and coordinating manual rotation with AI-controlled translation. 

\begin{table}[!h]
\caption{System Usability Scale (SUS) scores by \CamsWrittenOut{}}
\label{tab:SUS}
\centering
\scriptsize
\setlength{\tabcolsep}{3pt}
\renewcommand{\arraystretch}{1.05}
\begin{tabular}{llcccc}
\hline
 \# & \gls{cam}* & \textbf{Median} & Mean $\pm$ SD & Range & Grade \\
\hline
4 & Full Copilot &\textbf{77.5}& 73.4 $\pm$ 20.9& 22.5--100.0 & B\\
1 & Direct Robot Control & \textbf{75.0} & 74.8 $\pm$ 15.9 & 45.0--90.0 & B\\
3b & AI Rotation & \textbf{72.5}& 73.6 $\pm$ 14.4& 47.5--97.5 & B\\
2 & AI Advice & \textbf{57.5}& 53.4 $\pm$ 26.5& 17.5--92.5 & C \\
3a & AI Translation & \textbf{50.0}& 53.6 $\pm$ 21.3& 20.0--90.0 & C\\
\hline
\end{tabular}

\vspace{1mm}
\footnotesize{*From highest to lowest median SUS score. Standard grade ranges (A--F).}
\end{table}

\subsubsection{Ranking}\label{subsubsec_results_survey_ranking}
\tableref~\ref{tab:ranking} presents the participant preference rankings, ordered according to median rank.
\camnull{} was the most preferred condition overall, with the lowest mean rank ($1.64 \pm 1.29$) and the strongest first-choice frequency ($8/11$). 
Among the telerobotic and \AIdrivenmodes{}, \camone{} and \camfour{} were the most competitive alternatives with a mean rank of $3.00 \pm 1.18$ and $3.09 \pm 1.70$, respectively. Their preference patterns differed: \camone{} was ranked in the top three by $8/11$ clinicians. and no participant ranked it as the first or last choice; 
In contrast, preferences for \camfour{} were more variable, with $2/11$ participants ranking it first and $1/11$ ranking it last. Notably, these rankings at the extremes originated from consultant-level participants, suggesting greater heterogeneity in preferences within this subgroup.
The least favorable ranking overall showed \camthreea{} with a median rank of $6$ and ranked last by $6/11$ participants. 

\begin{table}[!t]
\caption{Ranking-based evaluation of \CamsWrittenOut{}}
\label{tab:ranking}
\centering
\scriptsize
\setlength{\tabcolsep}{3pt}
\renewcommand{\arraystretch}{1.05}
\begin{tabular}{llcccc}
\hline
\# & \gls{cam}* & \textbf{Median (IQR)} & Mean $\pm$ SD & 1st Choice & Last Choice \\ 
\hline 
0 & Manual & \textbf{1 [1.0--1.5]} & 1.64 $\pm$ 1.29 & 8/11 & 0/11 \\ 
1 & Direct Robot Control & \textbf{3 [2.0--3.5]} & 3.00 $\pm$ 1.18 & 0/11 & 0/11 \\ 
4 & Full Copilot & \textbf{3 [2.0--4.5]} & 3.09 $\pm$ 1.70 & 2/11 & 1/11 \\ 
2 & AI Advice & \textbf{4 [3.0--6.0]} & 4.18 $\pm$ 1.78 & 1/11 & 4/11 \\ 
3b & AI Rotation & \textbf{4 [4.0--4.5]} & 4.09 $\pm$ 0.70 & 0/11 & 0/11 \\ 
3a & AI Translation & \textbf{6 [4.5--6.0]} & 5.00 $\pm$ 1.41 & 0/11 & 6/11 \\
\hline
\end{tabular}

\vspace{1mm}
\footnotesize{*From lowest to highest median. Lower values indicate stronger preference.}
\end{table}

\section{Discussion}\label{sec_discussion}

\subsection{Interpretation of \CamsWrittenOut}\label{subsec_discussion_cams}

The evaluated \glspl{cam} revealed distinct trade-offs in the distribution of control authority between clinicians and the \gls{ai} agent. Increasing autonomy did not consistently improve objective performance or clinician acceptance. Instead, the perceived utility of each mode appeared to depend on whether autonomous behaviour was comprehensible, predictable and compatible with the clinician's ongoing control strategy.

\paragraph*{\textbf{AI Guidance}}
\camtwo{} received a moderate workload profile but a marginal \gls{sus} score, ranking second to last among the evaluated modes. Participants reported difficulty following the guidance while simultaneously monitoring fluoroscopic imaging and controlling the guidewire, describing it as rapidly changing, unclear, counterintuitive and affected by latency. This may reflect the agent's low-level control strategy, which generates repeated forward--backward and rotational movements to optimise guidewire pose before advancing. Presenting these instantaneous actions directly to the operator obscures the higher-level navigation strategy and may make the agent appear erratic. Several participants consequently appeared to reduce their reliance on the guidance after an initial attempt to follow it.

Nevertheless, a minority perceived the mode as beneficial, suggesting potential for the underlying concept with improved interaction design. Rather than streaming low-level actions, future systems could communicate higher-level navigation intentions and integrate them into the clinician's primary field of attention, for example by overlaying directional cues at the guidewire tip within the fluoroscopic image. Haptic guidance through the \gls{ucid}, such as \textit{Guided Control}~\cite{Yang2022_reviewHMC}, could provide an alternative communication channel with lower visual demand.

\paragraph*{\textbf{Shared Control}}
The two configurations of \camthree{}, \camthreea{} and \camthreeb{}, produced markedly different responses despite both distributing control authority between the clinician and the \agent{}. \camthreea{} received a marginal \gls{sus} score and the least favorable workload profile, with the highest ratings for effort, frustration and perceived performance demand. Its relatively short \gls{t2t} should be interpreted cautiously, as this metric included only successful trials and therefore excluded the substantial number of failures associated with its 38\% success rate.
The difficulties with \camthreea{} may be related to the disruption of the natural coupling between translation and rotation. Participants could not anticipate the \agent's actions, while the \agent{} could advance the guidewire before the operator had completed the required rotational adjustments. This reduced coordination between the two control dimensions.
In contrast, \camthreeb{} received a comparatively high \gls{sus} score and favorable workload profile. Participants reported that retaining direct robotic control over translation provided greater procedural influence and allowed them to slow or stop before critical bifurcations. Several also perceived the \agent{} as more effective at rotational manipulation than themselves. However, uncertainty regarding the \agent's intentions and coordination between translation and rotation remained concerns.

For both configurations, the \agent{} had been optimized primarily for autonomous navigation rather than clinician--AI collaboration. Its actions were therefore generated independently of operator intent, requiring clinicians to adapt to the \agent{} rather than the reverse. The contrasting responses to \camthreea{} and \camthreeb{} suggest that effective shared autonomy depends not only on the allocation of control authority but also on coordination between autonomous behavior and operator intent. Collaboration-aware policies that incorporate the clinician's ongoing behavior or inferred intention may therefore support more effective human--AI coordination.

\paragraph*{\textbf{User-Directed Delegation}}
\camfour{} achieved a favorable workload profile and one of the highest \gls{sus} scores when \camfoura{} and \camfourb{} were considered together, and received the most favorable overall ranking among the \AIdrivenmodes{}. Participants particularly valued \camfourb{}, which allowed them to invoke assistance selectively according to the anatomical context and task demands. Assistance could be used during challenging segments or during relatively straightforward anatomy to accelerate progression, with direct robot control resumed for critical steps or when performance deteriorated.

This flexibility did not eliminate concerns regarding autonomous performance. Participants reported anatomy-dependent failures, including unintended entry into the \gls{rsa}, failure to reach the target and additional recovery work when the \agent{} left the guidewire in an unfavorable position. The activation patterns provide further insight into how participants incorporated autonomous control. In \camfourb{}, assistance was most frequently activated near the beginning of the procedure, indicating early willingness to delegate irrespective of the target vessel.
Because bypassing the \gls{rsa} into the \gls{rcca} was particularly challenging, activation frequently occurred near this region following unsuccessful unaided attempts. Deactivation locations were more heterogeneous, consistent with participants disengaging the \agent{} when it selected an incorrect branch such as \gls{rica} when targeting the \gls{reca}, or \gls{rsa} when targeting the \gls{rcca}.

Together, these observations suggest that clinicians treated autonomous control as an opportunistic resource rather than a continuously active controller: they delegated control when autonomous behavior appeared beneficial and resumed direct robot control when it became unreliable. This preserved clinician authority while allowing selective use of AI assistance may explain the comparatively favorable perception of \camfour{} relative to \camtwo{} and \camthree{}. The findings further suggest that effective delegation requires not only reliable autonomous performance but also sufficient transparency for clinicians to anticipate when assistance is likely to be beneficial and recognize when it should be withdrawn. Communicating the agent's intended action or confidence could therefore support more informed delegation decisions.

\subsection{Implications for Clinician-AI Collaboration}
A key insight emerging from our results is that the shared-autonomy modes allowed compensation for limitations of the autonomous navigation agent.
\camfive{} achieved a success rate of only 23\%, yet several shared-autonomy modes built on the same underlying policy reached success rates of 92--97\%, comparable to or exceeding \camone{} (\secref{}~\ref{subsubsec_results_successrate}, \figref{}~\ref{fig:results_successrate}). This disparity suggests that overall performance depended less on the autonomous policy itself and more on how clinicians were able to supervise, interpret, or selectively deploy AI assistance. 
Thus, the limitation of sim-to-real domain shift shown here was largely mitigated.
The strongest example was \camfourb{}, which slightly exceeded the success rate of \camone{} despite relying on the same AI policy that performed poorly in isolation. This finding suggests that operator-directed task delegation may allow clinicians to exploit beneficial autonomous behaviors while intervening when AI actions become suboptimal. This interpretation is further supported by the \gls{t2t} results. 
Although \camfive{} exhibited the lowest success rate, completion times among successful trials were among the shortest observed (\secref{}~\ref{subsubsec_results_t2t}, \figref{}~\ref{fig:results_t2t}), indicating that the \agent{} possessed useful navigation competence when successful but lacked the robustness required for reliable autonomous operation.

Taken together, these findings indicate that an essential advantage of shared autonomy may not be to augment clinician performance directly, but to make imperfect AI systems clinically usable. 
Human oversight therefore appears to be more than a temporary safeguard until autonomous navigation becomes sufficiently reliable. Instead, clinician involvement may constitute an integral component of overall system performance.
Designing clinician-AI collaboration strategies that are robust to imperfect autonomy may conceivably prove more important for clinical translation than pursuing higher levels of autonomy alone.
Rather than representing an intermediate step toward full autonomy, shared-autonomy frameworks may constitute a long-term paradigm in which clinical expertise and AI capabilities complement one another, combining human robustness with the efficiency and task-specific strengths of autonomous systems.

\subsection{Study Limitations}\label{subsec_discussion_study_limitation} 
This study represents an initial investigation of clinician-AI collaboration in robotic neurointervention. Despite prospective power calculations, the small expert cohort ($10$ interventional neuroradiologists and $1$ interventional radiologist), substantial inter-operator variability, and modest differences between assistance modes limited the detection of statistically significant effects. Nevertheless, the recruited cohort represents almost 10\% of the UK's interventional neuroradiology workforce ($123$ in 2025)~\cite{RCR2025WorkforceCensus}, highlighting the practical challenges of conducting large-scale studies in this highly specialized field.

The relatively small sample size also limited the ability to investigate differences between participant subgroups. Whilst clinical experience was considered a potential confounder, our sensitivity analysis suggested that subgroup composition did not influence performance outcomes (\supmaterial{} A).

Furthermore, the generalizability of the findings is inherently limited, as the study examined a single navigation task within one patient-specific vascular anatomy. While the selected case contained clinically relevant challenges, it cannot represent the full diversity of neurointerventional procedures. Future work should therefore evaluate the proposed \glspl{cam} across a broader range of anatomies, navigation tasks, and clinical scenarios to determine whether the observed trends persist under varying procedural demands, where differences in workload, navigation complexity, and the value of AI assistance may become more pronounced.

Additional limitations arise from the experimental platform itself. The patient-specific phantom provided a controlled and reproducible environment for comparing \glspl{cam}, but did not capture physiological factors such as blood flow, vessel deformation, pulsatility, and realistic instrument--vessel interactions, all of which may influence navigation dynamics and affect both clinician and AI performance~\cite{Robertshaw2026positionstatement}. Similarly, participants reported perceived latency and reduced rotational precision when using the robotic control interface compared with manual manipulation. However, because the same robotic platform and interface were used across all robotic \glspl{cam}, it is plausible that these effects and further systematic errors are unlikely to account for the relative differences observed between assistance modes. In summary, while the comparative findings regarding clinician-AI collaboration are likely robust within the study setting, their translation to clinical practice should be validated in more realistic experimental and clinical environments.

\subsection{Sim-to-Real Gap}\label{subsec_discussion_limitations_agent}
Success rates decreased substantially from 81\% \insilico{} to 23\% \invitro{}, demonstrating a pronounced sim-to-real performance gap (\secref{}~\ref{subsubsec_results_successrate}). However, successful autonomous runs remained comparatively fast (\secref{}~\ref{subsubsec_results_t2t}, \figref{}~\ref{fig:results_t2t}), suggesting that locally effective navigation behaviors were preserved. One possible interpretation of these findings is that the principal challenge was not navigation efficiency itself, but maintaining robust performance over extended trajectories in the physical environment. 
This interpretation is consistent with the long-horizon, zero-shot nature of the investigated navigation task
(\secref{}~\ref{subsec_method_navigation_task}). 
Over such distances, small control inaccuracies, tracking errors, and sim-to-real mismatches can accumulate and ultimately result in procedural failure. Notably, several shared-autonomy modes achieved substantially higher success rates despite relying on the same underlying policy, suggesting that clinician supervision can compensate for a proportion of these failures. 
Nevertheless, improving sim-to-real transfer, state estimation, and long-horizon decision-making will likely be essential before fully autonomous robotic neurointerventions can be deployed reliably in clinical practice.

\section{Conclusion}\label{sec_conclusion}
This work introduced \CamsWrittenOut{}, a framework for systematically investigating clinician-AI collaboration in robotic endovascular navigation, and presented the first \invitrodash{} user study comparing multiple assistance paradigms across the autonomy spectrum with clinical experts. The implemented AI-driven modes did not outperform conventional robotic teleoperation, while hands-on manual navigation remained the strongest-performing mode overall. 
Nevertheless, several shared-autonomy approaches achieved performance comparable to robotic teleoperation despite limitations of the deployed \agent{}, plausibly suggesting that effective clinician-AI collaboration may depend less on increasing autonomy and more on the allocation of control authority between human operators and AI systems. 
Notably, all shared-autonomy modes substantially outperformed the \agent{} operating alone, indicating that certain clinician-AI collaboration strategies are robust to imperfect AI. 
Among the evaluated approaches, operator-directed task delegation and AI-assisted rotation showed the most favorable overall combination of objective performance, workload, usability, and clinician preference.
The study further provides one of the first empirical investigations of long-horizon autonomous endovascular navigation and highlights the substantial challenges associated with sim-to-real transfer. 
Overall, the findings indicate that shared-autonomy frameworks may have the potential to provide a practical pathway toward the clinical adoption of AI-assisted endovascular robotics. Future progress will likely depend on improving the robustness, generalizability, and clinical reliability of autonomous navigation systems while preserving meaningful clinician oversight.

\section*{Author Contribution}
S.M.H.S., C.B., and T.B. conceived the idea of a neuroendovascular copilot. N.F. designed the study, implemented the hardware and software components, performed pre-testing, conducted the experiments, analyzed the data, and wrote the majority of the manuscript. H.R. trained the \agent{} and contributed to manuscript writing. X.C. and W.D. supported experimental execution, data and survey analysis, and manuscript writing. B.J. contributed to experimental planning and developed the \gls{ucid}. S.M.H.S. contributed to the development of concepts for the \glspl{cam}, experimental planning, and the robotic manipulator. M.A.K., S.S., N.K., P.B., S.G., C.C., J.L., N.A., M.B., and A.F. provided clinical feedback during framework development and experimentation. A.G. contributed to manuscript revision and editing. C.B. and T.B. served as equal senior supervisors of the project and provided critical feedback and manuscript review throughout the study.
All authors read and approved the final manuscript.

\section*{Competing Interest}
All authors declare no financial or non-financial competing interests. 

\section*{Acknowledgments}
The authors thank Duane James, Gayathri Nantharatnam, and Marty Rajaratnam for their support in facilitating this study and the
operation of the mock operating room, as well as Raghav Khanna, Adam Warwicker, and Yi-Yi Liu for their help on manufacturing printed components, and Christoph Eyberg for providing the code base for automatic tool tracking.

Large Language Models (LLMs) were used to assist with the structure, grammar, and spelling of this manuscript and for the creation of visualizations. All outputs were reviewed, edited, and verified by the authors.

\bibliographystyle{IEEEtran}
\bibliography{references}

\end{document}